%% file: main.tex
\documentclass[10pt,twocolumn,letterpaper]{article}

\usepackage[accsupp]{axessibility}
\usepackage{cvpr}              

\input{preamble}

\definecolor{cvprblue}{rgb}{0.21,0.49,0.74}
\usepackage[pagebackref,breaklinks,colorlinks,citecolor=cvprblue]{hyperref}

\def\paperID{13009} 
\def\confName{CVPR}
\def\confYear{2024}

\title{Discriminative Sample-Guided and Parameter-Efficient Feature Space Adaptation for Cross-Domain Few-Shot Learning}

\author{Rashindrie Perera\textsuperscript{1,2} \qquad Saman Halgamuge\textsuperscript{1} 
\vspace{2mm}
\\
\textsuperscript{1} University of Melbourne, Australia \\
\textsuperscript{2} Peter MacCallum Cancer Centre, Australia \\ 
{\tt\small cdperera@student.unimelb.edu.au, saman@unimelb.edu.au}
}

\begin{document}
\maketitle
\input{sec/0_abstract}    
\input{sec/1_intro}
\input{sec/2_problem_definition}

\input{sec/3_method}

\input{sec/4_experiments}

\input{sec/5_results}

\input{sec/6_conclusions}
{
    \small
    \bibliographystyle{ieeenat_fullname}
    \bibliography{main}
}

\input{sec/x_supplementary}

\end{document}

%% file: preamble.tex
\usepackage[dvipsnames]{xcolor}


%% file: sec/0_abstract.tex
\begin{abstract}

In this paper, we look at cross-domain few-shot classification which presents the challenging task of learning new classes in previously unseen domains with few labelled examples. Existing methods, though somewhat effective, encounter several limitations, which we alleviate through two significant improvements. First, we introduce a lightweight parameter-efficient adaptation strategy to address overfitting associated with fine-tuning a large number of parameters on small datasets. This strategy employs a linear transformation of pre-trained features, significantly reducing the trainable parameter count. Second, we replace the traditional nearest centroid classifier with a discriminative sample-aware loss function, enhancing the model's sensitivity to the inter- and intra-class variances within the training set for improved clustering in feature space. Empirical evaluations on the Meta-Dataset benchmark showcase that our approach not only improves accuracy up to 7.7\% and 5.3\% on previously seen and unseen datasets, respectively, but also achieves the above performance while being at least $\sim3\times$ more parameter-efficient than existing methods, establishing a new state-of-the-art in cross-domain few-shot learning. Our code is available at \href{https://github.com/rashindrie/DIPA}{https://github.com/rashindrie/DIPA}.
\end{abstract}

%% file: sec/1_intro.tex
\section{Introduction}
\label{sec:intro}

Deep neural networks achieve remarkable performance when presented with abundant training data. However, collecting large datasets is not feasible in many applications due to bottlenecks in data (e.g., rare categories), or the cost of manual annotations. Inspired by this limitation, few-shot classification aims to learn a classifier to recognize new classes with only a limited number of samples per class \cite{NIPS2004_ef1e491a, 1597116, doi:10.1126/science.aab3050}. In traditional few-shot settings, new classes arise within a previously seen domain but have no class overlap with previously seen classes. Therefore, early works focused on learning to recognize new classes arising within a previously seen domain. However, in real-world scenarios, it is more likely that the new classes will arise from previously unseen domains. This challenging scenario of learning new classes in previously unseen domains is tackled in \textit{cross-domain} few-shot learning problem, which is also the focus of this paper.

Existing methods typically address this challenge by first learning a set of \textit{task-agnostic} (i.e., generalized) features from a large dataset. These features are then fine-tuned to a specific target task using a small training set (called \textit{support set}), often comprising as few as five images per category \cite{Triantafillou2020, Li2021Cross-domainAdapters}. These two stages are called meta-training (or pre-training) and meta-testing, respectively. The performance of the fine-tuned model is subsequently evaluated on a separate set of test samples, the \textit{query set}, where the objective is to accurately categorize each query sample into one of the classes represented in the support set. However, as illustrated in Table \ref{tab:param_counts}, existing methods frequently involve fine-tuning a considerable number of task-specific parameters during the meta-testing phase. This practice can lead to overfitting, especially when data is scarce \cite{Xu2023ExploringTransformers}. Moreover, it has been shown that shallower neural network layers generally contain more general features that can be applied to new tasks without the need for explicit fine-tuning \cite{Yosinski2014HowNetworks, zou2021revisiting}. Despite this, many existing studies \cite{Li2021Cross-domainAdapters, Hu2022PushingDifference, Xu2023ExploringTransformers} attempt to adapt the entire pre-trained model to the target task, which in a limited data regime, can increase the risk of overfitting.

\begin{table}[!htbp]
\centering
\resizebox{\columnwidth}{!}{%
\begin{tabular}{l|c|c|c|c}
\hline

Model & Backbone & \# Total  & \# Fine-tuned  & \% Fine-tuned  \\
& & Params & Params & Params \\
& & (T) & (F) & (F/T)*100 \\ \hline

TSA \cite{Li2021Cross-domainAdapters}, TSA+SSA \cite{SreenivasSimilarLearning} & ResNet-18 & 12.65 M & 1.48 M & 11.7\% \\
PMF \cite{Hu2022PushingDifference} & ViT-Sma1l & 21.72M & 21.72M & 100.0\% \\
eTT \cite{Xu2023ExploringTransformers} & ViT-Sma11 & 21.72M & 1.95 M & 9.0\% \\
ATTNSCALE \cite{Basu2023StrongBaseline} & ViT-Sma11 & 21.72M & 0.26 M & 1.2\% \\ \hline
Ours, $d_{t}$ = 12 & ViT-Sma11 & 21.72 M & 0.10 M & 0.5\% \textcolor{green}{($\downarrow$ 0.7)} \\
Ours, $d_{t}$ = 9 & ViT-Sma11 & 21.72M & \textbf{0.08 M} & \textbf{0.4\%} \textcolor{green}{($\downarrow$ 0.8)} \\ 
Ours, $d_{t}$ = 7 & ViT-Sma11 & 21.72M & \textbf{0.06 M} & \textbf{0.3\%} \textcolor{green}{($\downarrow$ 0.9)} \\ \hline
\end{tabular}%
}
\caption{Comparison of the number of parameters fine-tuned during the meta-testing stage by recent state-of-the-art methods, including our method (last two rows). For our method, $d_t$ refers to tuning depth, where $d_t=12$ fine-tunes every layer of the model while $d_t=7$ fine-tunes only the last 7 layers of the model.}
\label{tab:param_counts}
\end{table}

\begin{figure}
    \centering
    \includegraphics[width=1\linewidth]{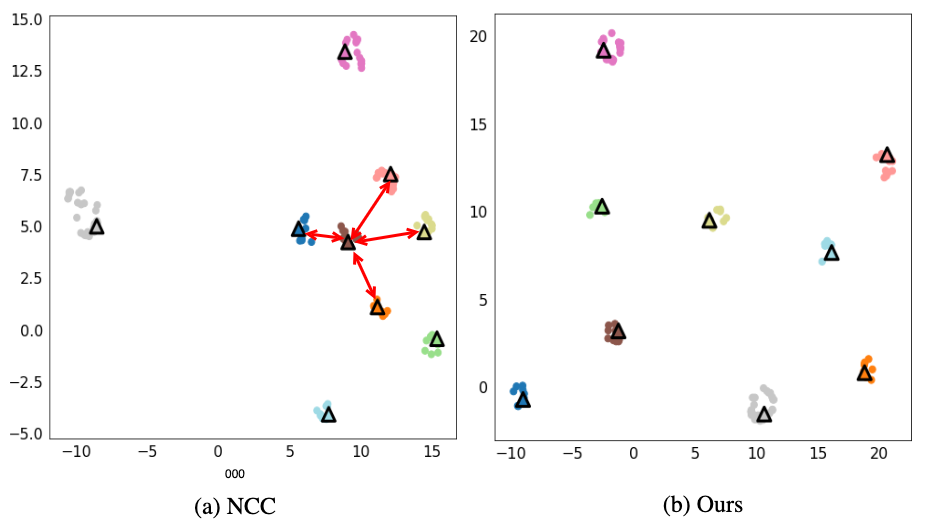}
    \caption{Visualization of feature space adaptation using the support set by employing (a) NCC and (b) Our approach for a few-shot classification task on ImageNet \cite{russakovsky2015imagenet}. The $\Delta$ represents class centroids. The clusters formed by NCC are located close to each other, thereby, potentially generating confusing class centroids. In contrast, clusters formed in our approach are well separated.}
    \label{fig:cluster_vis_training}
\end{figure}

On the other hand, existing works \cite{Li2021Cross-domainAdapters, QinBi-levelGeneralization, SreenivasSimilarLearning, Triantafillou2020} typically employ the nearest centroid classifier (NCC) \cite{mensink2013distance, snell2017prototypical} for fine-tuning the task-specific parameters and for subsequent query classification. NCC assigns an image to the class with the closest centroid, where the centroid is the mean of the feature embeddings belonging to the class. It encourages each embedding to be closer to its respective class centroid than the centroids of the other classes. As a result, these classes form tight clusters in the feature embedding space (refer Fig. \ref{fig:cluster_vis_training}a). However, due to the lack of focus on inter-class variance in NCC \cite{mensink2013distance}, there is a risk that the resulting clusters may not be adequately separated. This can lead to centroids being positioned too closely, as depicted in Fig. \ref{fig:cluster_vis_training}a, causing confusion during query classification.

Motivated by the above, as shown in Fig. \ref{fig:fsl_pipeline}, we propose to employ a set of lightweight task-specific parameters for adapting the task-agnostic feature backbone to an unseen domain. Unlike prior work \cite{Hu2022PushingDifference, Xu2023ExploringTransformers, Li2021Cross-domainAdapters, Basu2023StrongBaseline}, we propose to employ a parameter-efficient linear transformation of features for the adaptation of a pre-trained model which significantly reduces the count of trainable parameters. Also, we propose to vary the tuning depth to fine-tune only the less transferable feature representations in the deeper layers of the network, thereby further reducing the number of trainable parameters, making it $3.3\times$ to $4.3\times$ more parameter-efficient than existing methods (refer Table \ref{tab:param_counts} where $d_t=9, 7$). Moreover, we propose to utilize the sample information present within the support set to identify challenging positive examples (i.e., the most dissimilar positive examples of a given class) or challenging negative examples (i.e., the most similar negative examples for a given class) for encouraging better intra and inter-class separation during fine-tuning to encourage better cluster formation in the feature space. Consequently, as shown in Fig. \ref{fig:cluster_vis_training}b, our approach forms well-separated clusters in the feature space, minimizing confusing centroids and thereby enhancing query classification accuracy. We advance few-shot learning with a straightforward and effective pipeline. We systematically evaluate our method on the standard cross-domain few-shot classification benchmark and demonstrate superior performance to all well-known methods.

%% file: sec/2_problem_definition.tex
\section{Related Work}
\label{sec:notation}

Below we provide a more detailed discussion of the most related work. 

Learning task-agnostic representations that are easily adaptable to new tasks is crucial for cross-domain generalization success \cite{bilen2017universal, Li2021Cross-domainAdapters}. Many existing methods \cite{snell2017prototypical, Bateni2022BeyondLearning, Triantafillou2021LearningGeneralization, doersch2020crosstransformers, Li2021Cross-domainAdapters, SreenivasSimilarLearning, QinBi-levelGeneralization} rely on supervised learning during meta-training to learn task-agnostic representations, but this approach can lead to \textit{supervision collapse} \cite{Doersch2020, doersch2020crosstransformers}, where models only focus on features important for the source dataset’s classes and thereby ignore semantically important features of unseen classes. To circumvent this, our approach employs self-supervised learning for pre-training a task-agnostic feature extractor, ensuring broader learning beyond the source dataset’s class labels \cite{Hu2022PushingDifference, Xu2023ExploringTransformers, Basu2023StrongBaseline}

During meta-testing, task-agnostic representations are adapted (or fine-tuned) to the target task using the support set. Simple CNAPS \cite{Bateni2022BeyondLearning} and FLUTE \cite{Triantafillou2021LearningGeneralization} employ task-specific FiLM layers \cite{Perez2018FiLM:Layer}, which are connected serially to the backbone and involve affine transformations for feature extractor adaptation, with FiLM parameters estimated by a meta-trained auxiliary network. Unlike them, our method directly learns this adaptation on the support set, eliminating the need for auxiliary networks. TSA \cite{Li2021Cross-domainAdapters} adapts the full backbone of a pre-trained extractor using residual adapters, which are connected in parallel and involve matrix multiplications \cite{rebuffi2018efficient}. Similarly, eTT \cite{Xu2023ExploringTransformers} and ATTNSCALE \cite{Basu2023StrongBaseline} utilize visual prompts or scaling matrices for task-specific adaptations. In contrast, our method simplifies this process by using lightweight, linear transformations for adapting features to the target task. URT \cite{Liu2020AClassification} uses multiple domain-specific feature extractors with a task-specific fusion mechanism, which increases training costs. In contrast, we achieve adaptation with a single feature extractor and task-specific parameters. Additionally, rather than adapting the entire feature backbone like many existing works \cite{Li2021Cross-domainAdapters, Hu2022PushingDifference, Basu2023StrongBaseline, Xu2023ExploringTransformers}, we focus on adapting the deeper network layers. Finally, we diverge from the common use of NCC loss for the fine-tuning process in prior studies \cite{snell2017prototypical, Triantafillou2020, Li2021Cross-domainAdapters, Hu2022PushingDifference, SreenivasSimilarLearning, Basu2023StrongBaseline, Xu2023ExploringTransformers} by employing a discriminative sample-aware loss function on the support set.

There are multiple existing works that do not fall under the typical task-agnostic and task-specific categorization discussed above. For instance, ProtoNet \cite{snell2017prototypical} utilizes class prototypes to classify query samples. CTX \cite{doersch2020crosstransformers} extends ProtoNet by using attention mechanisms to create more task-aligned prototypes. SSA \cite{SreenivasSimilarLearning} improves upon TSA by augmenting the support set dataset to create more challenging training examples. PMF \cite{Hu2022PushingDifference} fine-tunes the full feature backbone for the target domain with an added task sampling and learning rate selection strategy. Our approach is also benchmarked against these for a comprehensive performance comparison.

%% file: sec/3_method.tex
\section{Methodology}
\label{sec:method}


\begin{figure*}[ht]
  \centering
   \includegraphics[width=\textwidth]{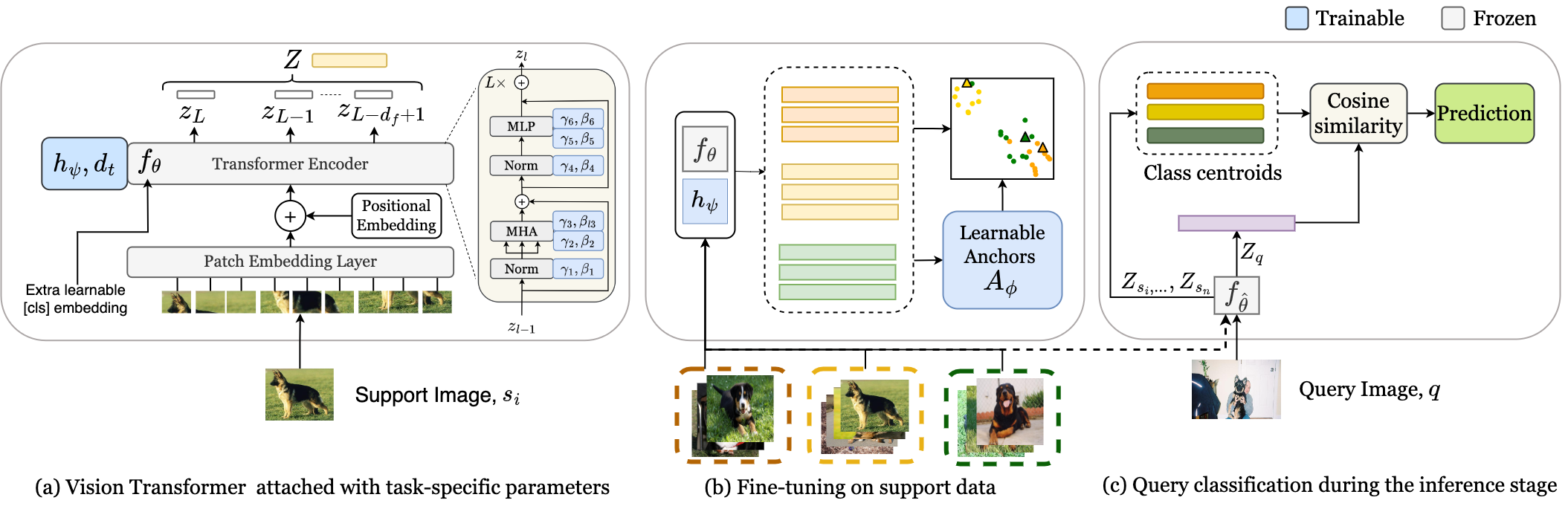}
   \caption{Illustration of our framework during meta-testing. \textbf{(a)}: A set of task-specific parameters $ (\gamma_m, \beta_m) \in h_{\psi}$ are attached to a ViT backbone $f_{\theta}$, up to a pre-defined tuning depth $d_t$, where $m=1,..,6$. \textbf{(b)} $h_{\psi}$ is fine-tuned on the support set using a set of learnable anchors $A_{\phi}$. \textbf{(c)} Query images are classified by assigning them to the nearest class centre using the fine-tuned model $f_{\hat{\theta}}$. $L$: Number of layers in the ViT, $d_f$: feature fusion depth, $z_l$: $[cls]$ output from $l$\textsuperscript{th} ViT layer, $Z$: fused feature embedding, Norm: Layer Normalization, MLP: Multi-Layer Perceptron, and MHA: Multi-head Attention.}
   \label{fig:fsl_pipeline}
\end{figure*}


A few-shot task typically contains support set $\mathcal{S}= \{x_i, y_i \}_{i=1}^{|S|}$ containing $|S|$ sample and label pairs and a query set $\mathcal{Q}= \{x_i, y_i\}_{i=1}^{|Q|}$ containing $|Q|$ sample and label pairs. The goal in few-shot classification is to learn a classifier using $\mathcal{S}$ that accurately predicts the labels of $\mathcal{Q}$. Note that this paper focuses on the few-shot image classification problem, i.e. $x$ and $y$ denote an image and its label.

We follow existing work \cite{Dvornik2020SelectingClassification, Li2021UniversalClassification, Xu2023ExploringTransformers} and employ a two-step approach to solve this problem. In the first stage, we train a feature extractor $f_\theta$ to learn task-agnostic feature representations using a large dataset (called source dataset). In the second stage, we first adapt the task-agnostic features for a target task \((\mathcal{S}, \mathcal{Q})\) sampled from the target dataset, using task-specific weights learned from the support set \(\mathcal{S}\). Subsequently, we use the adapted model to classify the query samples in \(\mathcal{Q}\).

\textbf{\textit{Neural architecture}}: As in \cite{Hu2022PushingDifference, Xu2023ExploringTransformers, Basu2023StrongBaseline}, we employ a Vision Transformer (ViT) model as our backbone architecture.

\subsection{Task-agnostic representation learning}

Recent studies \cite{Hu2022PushingDifference, Xu2023ExploringTransformers, Basu2023StrongBaseline} propose utilizing the DINO self-supervised algorithm \cite{Caron2021EmergingTransformers} to pre-train the task-agnostic feature representations. DINO focuses on learning from the `global-to-local' relationship within image crops to derive insightful features. Our approach, however, draws on the emerging concept of masked image modelling (MIM) \cite{Zhou2021IBOT:Tokenizer, he2022masked, devlin2018bert, Hiller2022RethinkingClassification} for pre-training the feature extractor $f_\theta$. MIM involves masking certain patches of an image and prompting the model to reconstruct these masked portions utilizing the context of the unmasked patches. This method requires the model to infer not just the missing visual information but also the context in which it occurs, necessitating a deep understanding of the image content, thereby, facilitating the learning of semantically rich and generalized feature representations. As illustrated in our results, this generalization is key for learning task-agnostic feature representations that can easily adapt to new tasks.

\subsection{Task-specific representation learning}

The feature extractor $f_{\theta}$ trained using MIM is expected to provide a good starting point for more advanced image recognition tasks \cite{Zhou2021IBOT:Tokenizer}. However, when dealing with new classes from unseen domains, these features often require fine-tuning to better handle the novel scenarios.

To adapt $f_{\theta}$ to the target task, we propose to attach a set of scaling and shifting offset parameters to the pre-trained ViT backbone to learn a linear transformation of features \cite{Lian2022ScalingTuning}. As discussed above, unlike existing works that employ matrix-based \cite{Li2021Cross-domainAdapters, Basu2023StrongBaseline} or prompt-based \cite{Xu2023ExploringTransformers} calculations, we strictly apply these offsets as a linear transformation. Specifically, we only learn the amount of scale ($\gamma$) and shift ($\beta$) required for adapting the task-agnostic features to the target task, thereby, reducing the tunable parameter count to only $0.5\%$ of the model parameters during meta-testing. Formally, in a ViT, for an input $x \in \mathbb{R}^{(P^2+1) \times e}$, where $e$ is the embedding dimension, the output $y \in \mathbb{R}^{(P^2+1) \times e}$ (which is also the input for the next layer) is computed as, 

\begin{equation}
     y = \gamma \odot x + \beta
\end{equation} 

where $\gamma \in \mathbb{R} $ and $\beta \in \mathbb{R}$ are the amount of scale and shift applied on $x$, and $\odot$ is the dot product. This approach draws intuition from the concept of feature distribution matching \cite{sun2016return, Lian2022ScalingTuning}. Specifically, the aim is to tune the first-order (mean) and second-order (variance) statistics of feature distributions, enabling adjustment of pre-trained features to the target data distribution using only two parameters per layer.

We apply this linear transformation on the layer normalization, multi-layer perceptron, and multi-head self-attention layers of the ViT as illustrated in Fig. \ref{fig:fsl_pipeline}a. While the choice of tuned layers can be varied, we opt for our experiments to tune all three types of layers. We denote the final adapted feature extractor that includes the re-parameterized task-specific parameters as $f_{\hat{\theta}}$.

\textbf{Varying the tuning depth}. As discussed above, recent works \cite{Li2021Cross-domainAdapters, Hu2022PushingDifference, Xu2023ExploringTransformers, Basu2023StrongBaseline} commonly fine-tune the full backbone for the novel task. Instead, our method varies the extent of task-specific adaption of a pre-trained model for the novel task. This approach enables customized adaptation to new tasks by selectively fine-tuning pre-trained representations across the model’s layers to meet the specific requirements of each task. Specifically, we vary the depth $d_t$ of task-specific parameters $h_{\psi}$ attached to the pre-trained model (i.e., the depth of layers that are fine-tuned on the target task) where $0 < d_t < L$, and $h_{\psi} = \{\psi_{j}$, where $j=(L-d_t+1), ..., 12\}$ and $\psi_{j} = \{(\gamma_{m}, \beta_{m})$, where $m=1,..,6\}$. This approach, which strategically reduces the number of tunable parameters during the meta-testing phase as detailed in Table \ref{tab:param_counts}, is designed to further mitigate the risk of overfitting on novel tasks.

\subsubsection{Discriminative sample-guided feature adaptation}
\label{sec:variance}

We aim to provide discriminative sample-guided supervisory signals to the model, considering the relations between the samples in the support set. For this, we propose to utilize a discriminative sample-aware loss function. Similar to NCC, we assign a learnable class representative (prototype) for each class in the support set. However, unlike NCC, we treat these prototypes as \textit{anchors} (denoted by $A_{\phi}$) to which the samples can be attracted or repelled to learn a task-adapted embedding space. Moreover, unlike NCC which only associates the samples of a given class to its class prototype, we propose to utilize the anchors to associate with all the samples in the support set. This can be denoted as,

\begin{multline} 
l_{A_\phi}(X) = \frac{1}{|A|} \sum_{a \in A} \biggl\{ log (1 + \sum_{x \in X^+_a} e^{ \alpha(\delta - s(x, a))}) \\
+  log (1 + \sum_{x \in X^-_a} e^{\alpha(s(x, a) + \delta))}  \biggl\}\
\label{eq:pa_loss}
\end{multline}

where for a set of embedding vectors $X$ and an anchor $a \in A$, the loss encourages the cosine similarity $s(x, a)$ between a feature vector $x$ $\in$ $X$ and $a$ to be larger (i.e., greater than a user-defined margin $\delta$), if $a$ corresponds to the anchor of the class $x$ belongs to (denoted as x $\in$ $X^+_a$), or smaller (i.e., less than $-\delta$) if $x$ belongs to a different class (denoted as x $\in$ $X^-_a$). $\alpha > 0$ is a user-defined scaling factor. $l_{A_\phi}(X)$ is known as the proxy anchor loss \cite{Kim2020ProxyLearning}.

\begin{figure}[ht]
  \centering
   \includegraphics[width=1\columnwidth]{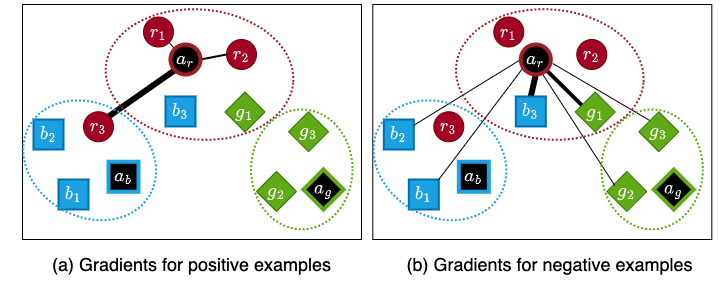}

   \caption{Loss gradient calculation during training. The example illustrates a scenario with three unique classes red, green and blue, denoted by $r, g, b$. The anchor of each class is coloured in black and denoted as $a_{class}$. \textbf{(a)} The gradients for positive samples of class $r$ are computed based on the relative hardness of all positive samples, so as to pull harder positives more strongly (thicker black lines). \textbf{(b)} The gradient calculation for negative samples for class $r$ considers the distribution of all negative samples and tries to push harder negatives more strongly (thicker black lines).}
   \label{fig:pa_gradient}
\end{figure}

Specific to our work is the manner in which the loss modifies the feature space by considering the sample information present within the support set, as illustrated in Fig. \ref{fig:pa_gradient}. Here, for a given anchor $a \in A$, it tries to pull $a$ and its most challenging positive example (for instance, $a_r$ and $r_3$ in Fig. \ref{fig:pa_gradient}a) together. Similarly, it tries to push $a$ and its most challenging negative example (for instance, $a_r$ and $b_3$ in Fig. \ref{fig:pa_gradient}b) apart. The gradient is larger (thicker lines) when the positive feature vector is far from $a$, and when the negative feature vector is close to $a$. In this manner, Eq. \ref{eq:pa_loss} considers the \textit{relative difficulty (hardness)} of each sample based on the inter and intra-class variation present in the feature embedding space to determine the relative strength of pull and push force to be applied for each sample. As demonstrated in our results, such variance-guided gradients provide better supervisory signals for fine-tuning in comparison to the conventional NCC.

\subsection{Feature fusion}

Features from shallower layers of the network have been observed to be more transferrable to unseen domains than those from deeper layers \cite{Yosinski2014HowNetworks, Caron2021EmergingTransformers}. Yet, existing works \cite{Li2021Cross-domainAdapters, Basu2023StrongBaseline, Bilen2017UniversalBreeds, Hu2022PushingDifference, Basu2023StrongBaseline, Xu2023ExploringTransformers} only use the features from the last layer of the model to represent an image, i.e. feature depth $d_f=1$. In contrast to them, we extract the $[cls]$ embedding from the last $d_f > 1$ layers of the ViT and concatenate them to form the final feature embedding for an input image as,
\begin{equation} 
Z = concat(z_{{L}}, z_{{L-1}}, ..., z_{{L-d_f+1}})
\label{eq:feat_fusion}
\end{equation}

where $z_{{L}}$ is the $[cls]$ embedding output for an image from the $L$\textsuperscript{th} layer, $d_f > 1$ is a hyper-parameter, $z\in\mathbb{R}^{dim}$ and $Z\in\mathbb{R}^{h*dim}$. Although this straightforward fusion strategy requires no additional training as some existing feature fusion strategies \cite{Vinyals2016, Zhang2020DeepEMD:Learning, li2023deep}, we observe significant performance gains when we increase the fusion depth.

\subsection{Query classification}

Following the feature space adaptation, we combine the adapted feature extractor \(f_{\hat{\theta}}\) with a NCC classifier to classify query samples. In NCC, a prototype \(c_n\) is calculated for each class \(n\) in the support set as the mean of all feature representations belonging to that class, defined as:

\begin{equation} 
c_n = \frac{1}{|S_n|} \sum\limits_{(x, y) \in S_n} f_{\hat{\theta}}(x), \quad n=1,..., N 
\end{equation}

Here, \((x, y) \in S_n\) represents pairs of feature embeddings and their corresponding labels in the support class \(S_n\). To classify a sample $x_q$, we employ cosine similarity as the distance metric $d(f_{\hat{\theta}}(x_q), c_{n})$ to assign \(x_q\) to the class with the nearest centroid \(c_n\) \cite{snell2017prototypical, Li2021Cross-domainAdapters}.

%% file: sec/4_experiments.tex
\section{Experiments}

\subsection{Experimental Settings}

\textbf{Dataset:} We use the standard cross-domain benchmark dataset, Meta-Dataset \cite{Triantafillou2020}. It contains images from 13 diverse datasets. We follow the standard protocol \cite{Triantafillou2020} for 1) multi-domain learning (MDL), where the train sets of eight datasets from Meta-Dataset are used for pre-training and 2) single-domain learning (SDL), where the train set of only ImageNet is used for pre-training. For fair comparisons, we also follow \cite{Hu2022PushingDifference} for single-domain learning with extra data (SDL-E) setting, where the entire ImageNet dataset is used for pre-training. Refer to Supplementary for more details.

\textbf{Evaluation}: We sample 600 $N$-way-$K$-shot tasks from the test set of each dataset and report the average accuracy and 95\% confidence score. Here, $N$ denotes the number of classes and $K$ denotes the number of examples per class present in the task where the convention is to sample the $N$, $K$ and the number of query images uniformly at random (refer \cite{Triantafillou2020} Appendix.3 for more details). We evaluate the Meta-Dataset on MDL, SDL and SDL-E settings.

\textbf{Architecture}: We limit our experiments to the standard ViT-small architecture \cite{dosovitskiy2020image}. Recent ViT-based works \cite{Hu2022PushingDifference, Xu2023ExploringTransformers}, including ours, typically resize input images to $224 \times 224$ pixels, deviating from earlier approaches that use $84 \times 84$ pixels \cite{Li2021Cross-domainAdapters, Triantafillou2021LearningGeneralization}. We acknowledge that higher resolution may impact classification performance due to richer visual information. Therefore, we include backbone architecture details in our results for transparent method comparison.

\textbf{Pre-training}: We follow the strategy proposed by \cite{Zhou2021IBOT:Tokenizer} and mostly stick to the hyperparameter settings reported in their work. A batch size of 128, 800 epochs and 4 Nvidia A100 GPUs with 80GB each is used for pre-training. We refer to Supplementary for more details.

\textbf{Fine-tuning}: We experimentally determine the values for the hyperparamters, where $\delta=0.1$, $\alpha=32$, and $d_f=4$ give optimal performance across domains. $\gamma$ and $\beta$ parameters were initialized with constant values of one and zero, i.e., without any randomness, ensuring the reproducibility of our results. We use two separate NAdam optimizers \cite{dozat2016nadam} with the learning rate of $0.005$ for $h_{\psi}$ and $5$ for $A_{\phi}$. We experimentally find that this learning rate combination works well across most tasks. Fine-tuning is performed across $80$ iterations on a single Nvidia A100 GPU with 80GB. We refer to Supplementary Section \ref{appendix:Hyperparameters}  for more details on hyperparameters. Finally, we report results using $d_t=7$ for domains seen during pre-training (in-domain) and $d_t=9$ for previously unseen domains (out-of-domain) as they provided the optimal average results across domains. Section \ref{sec:additional_analysis} provides more discussion on this choice.

%% file: sec/5_results.tex
\section{Main Results}

\subsection{Comparison to state-of-the-art methods}

\begin{table*}[!htbp]
\centering
\resizebox{\textwidth}{!}{%
\begin{tabular}{lc|c|c|c|ccccccccccccccccc}
\hline
ID & \multicolumn{1}{l|}{Method} & \multicolumn{1}{c|}{SS} &  \multicolumn{1}{c|}{Sup.} & \multicolumn{1}{c|}{Back} & ImageNet & Omniglot & Aircraft & Birds & Textures & QuickDraw & Fungi & \multicolumn{1}{c|}{VGGFlower} & TrafficSign & MSCOCO & MNIST & CIFAR10 & \multicolumn{1}{c|}{CIFAR100} & Avg. &  Avg. & Avg. &  Avg. \\ 
 & \multicolumn{1}{l|}{} &\multicolumn{1}{l|}{PT} & \multicolumn{1}{l|}{MT} & \multicolumn{1}{c|}{-bone} & \multicolumn{8}{c|}{} & \multicolumn{5}{c|}{} & Seen & Unseen & Unseen & All \\ \cline{6-18}
 & & & \multicolumn{1}{l|}{} & \multicolumn{1}{c|}{} & \multicolumn{8}{c|}{In-domain} & \multicolumn{5}{c|}{Out-of-domain} &  & (com.) & (all) &  \\ \hline

& \multicolumn{1}{l|}{\textbf{PT/MT=MDL}} & & &\multicolumn{1}{l|}{} & \multicolumn{1}{l}{} & \multicolumn{1}{l}{} & \multicolumn{1}{l}{} & \multicolumn{1}{l}{} & \multicolumn{1}{l}{} & \multicolumn{1}{l}{} & \multicolumn{1}{l}{} & \multicolumn{1}{l|}{} & \multicolumn{1}{l}{} & \multicolumn{1}{l}{} & \multicolumn{1}{l}{} & \multicolumn{1}{l}{} & \multicolumn{1}{l|}{} & \multicolumn{1}{l}{} & \multicolumn{1}{l}{} &  \multicolumn{1}{l}{} \\

A0 & \multicolumn{1}{l|}{Simple CNAPS \cite{Bateni2022BeyondLearning}}& &$\checkmark$& \multicolumn{1}{c|}{RN18} & 58.4 ± 1.1 & 91.6 ± 0.6 & 82.0 ± 0.7 & 74.8 ± 0.9 & 68.8 ± 0.9 & 76.5 ± 0.8 & 46.6 ± 1.0 & \multicolumn{1}{c|}{90.5 ± 0.5} & 57.2 ± 1.0 & 48.9 ± 1.1 & 94.6 ± 0.4 & 74.9 ± 0.7 & \multicolumn{1}{c|}{61.3 ± 1.1} & 73.7 &  53.1 & 67.4 & 71.2 \\
A1 & \multicolumn{1}{l|}{URT \cite{Liu2020AClassification}}&& $\checkmark$& \multicolumn{1}{c|}{RN18}  & 56.8 ± 1.1 & 94.2 ± 0.4 & 85.8 ± 0.5 & 76.2 ± 0.8 & 71.6 ± 0.7 & 82.4 ± 0.6 & 64.0 ± 1.0 & \multicolumn{1}{c|}{87.9 ± 0.6} & 48.2 ± 1.1 & 51.5 ± 1.1 & 90.6 ± 0.5 & 67.0 ± 0.8 & \multicolumn{1}{c|}{57.3 ± 1.0} & 77.4 & 49.9 & 62.9 & 71.8 \\
A2& \multicolumn{1}{l|}{FLUTE \cite{Triantafillou2021LearningGeneralization}}&  & $\checkmark$&\multicolumn{1}{c|}{RN18} & 58.6 ± 1.0 & 92.0 ± 0.6 & 82.8 ± 0.7 & 75.3 ± 0.8 & 71.2 ± 0.8 & 77.3 ± 0.7 & 48.5 ± 1.0 & \multicolumn{1}{c|}{90.5 ± 0.5} & 63.0 ± 1.0 & 52.8 ± 1.1 & 96.2 ± 0.3 & 75.4 ± 0.8 & \multicolumn{1}{c|}{62.0 ± 1.0} & 74.5 & 57.9  & 69.9 & 72.7 \\
A3& \multicolumn{1}{l|}{TSA \cite{Li2021Cross-domainAdapters}} & &$\checkmark$& \multicolumn{1}{c|}{RN18} & 59.5 ± 1.0 & 94.9 ± 0.4 & 89.9 ± 0.4 & 81.1 ± 0.8 & 77.5 ± 0.7 & 81.7 ± 0.6 & 66.3 ± 0.8 & \multicolumn{1}{c|}{92.2 ± 0.5} & 82.8 ± 1.0 & 57.6 ± 1.0 & 96.7 ± 0.4 & 82.9 ± 0.7 & \multicolumn{1}{c|}{70.4 ± 0.9} & 80.4 & 70.2 &  78.1 & 79.5 \\
A4& \multicolumn{1}{l|}{TSA + SSA \cite{SreenivasSimilarLearning}} & &$\checkmark$& \multicolumn{1}{c|}{RN18} & 58.9 ± 1.1 & \textbf{95.6 ± 0.4} & \textbf{90.0 ± 0.5} & 82.2 ± 0.7 & 77.6 ± 0.7 & \textbf{82.7 ± 0.7} & 66.6 ± 0.8 & \multicolumn{1}{c|}{93.0 ± 0.5} & 84.9 ± 1.1 & 58.1 ± 1.0 & \textbf{98.5 ± 0.4} & 82.9 ± 0.7 & \multicolumn{1}{c|}{70.8 ± 0.9} & 80.8 & 71.5  & 79.0 & 80.1 \\
A5& \multicolumn{1}{l|}{eTT \cite{Xu2023ExploringTransformers}} &$\checkmark$ & &\multicolumn{1}{c|}{ViT-s} & 67.4 ± 1.0 & 78.1 ± 1.2 & 79.9 ± 1.1 & 85.9 ± 0.9 & 87.6 ± 0.6 & 71.3  ± 0.9 & 61.8 ± 1.1 & \multicolumn{1}{c|}{96.6 ± 0.5} & 85.1 ± 0.9 & 62.3 ± 1.0 &  &  & \multicolumn{1}{c|}{} & 79.6 & 73.7  & 73.7  &  78.3 \\
A6& \multicolumn{1}{l|}{DIPA} &$\checkmark$ & &\multicolumn{1}{c|}{ViT-s} & \textbf{70.9 ± 1.0} & 84.7 ± 1.1 & 86.3 ± 1.0 & \textbf{90.8 ± 0.8} & \textbf{88.6 ± 0.5} & 75.3 ± 0.8 & \textbf{66.6 ± 1.1} & \multicolumn{1}{c|}{\textbf{97.9 ± 0.3}} & \textbf{91.3 ± 1.0} & \textbf{64.8 ± 1.0} & 96.9 ± 0.5 & \textbf{87.4 ± 0.6} & \multicolumn{1}{c|}{\textbf{81.2 ± 0.8}} & \textbf{82.6}  & \textbf{78.1} & \textbf{84.3} & \textbf{83.3} \\ 
& && \multicolumn{1}{l|}{}  & \multicolumn{1}{l|}{} &  &  &  & &  & & & \multicolumn{1}{c|}{} &  &  & &  & \multicolumn{1}{c|}{} & \textbf{\textcolor{green}{(+1.8)}}  & \textbf{\textcolor{green}{(+4.4)}} & \textbf{\textcolor{green}{(+5.3)}} & \textbf{\textcolor{green}{(+3.2)}} \\ 

& \multicolumn{1}{l|}{\textbf{\begin{tabular}[c]{@{}l@{}}PT=SDL-E, MT=MDL \end{tabular}}}  & &\multicolumn{1}{l|}{}  &  &&  &  & \textbf{} & \textbf{} &  &  & \multicolumn{1}{c|}{\textbf{}} & \textbf{} & \textbf{} &  &  & \multicolumn{1}{c|}{} &  &  &  \\ 
B0 & \multicolumn{1}{l|}{PMF \cite{Hu2022PushingDifference}} & $\checkmark$& $\checkmark$&\multicolumn{1}{c|}{ViT-s} & \multicolumn{1}{c}{74.6} & \textbf{91.8} & \textbf{88.3} & 91.0 & 86.6 & \textbf{79.2} & \textbf{74.2} & \multicolumn{1}{c|}{94.1} & 88.9 & 62.6 &  &  & \multicolumn{1}{c|}{} & \textbf{85.0} & 75.8 & 75.8 & 83.1 \\
B1& \multicolumn{1}{l|}{ATTNSCALE \cite{Basu2023StrongBaseline}}& $\checkmark$ & &\multicolumn{1}{c|}{ViT-s}  & \multicolumn{1}{c}{\textbf{}} & 80.9 & 78.8 & 86.7 & 85.8 & 74.4 & 59.0 & \multicolumn{1}{c|}{95.9} & 91.4 & 61.0 &  &  & \multicolumn{1}{c|}{} & 80.2 & 76.2 & 76.2 & 79.3 \\

B2& \multicolumn{1}{l|}{DIPA} & $\checkmark$ & &\multicolumn{1}{c|}{ViT-s} & \multicolumn{1}{c}{\textbf{77.3 ± 0.7}} & 83.9 ± 1.1 & 86.0 ± 1.1 & \textbf{91.1 ± 0.7} & \textbf{88.8 ± 0.5} & 75.9 ± 0.8 & 62.4 ± 1.1 & \multicolumn{1}{c|}{\textbf{97.7 ± 0.3}} & \textbf{91.7 ± 0.8} & \textbf{66.5 ± 0.9} & \textbf{97.2 ± 0.5} & \textbf{92.2 ± 0.5} & \multicolumn{1}{c|}{\textbf{84.5 ± 0.7}} & 82.9  & \textbf{79.1} & \textbf{86.4} & \textbf{84.2} \\ 
& \multicolumn{1}{l|}{} & & \multicolumn{1}{l|}{} &  &  & & & &  & & & \multicolumn{1}{c|}{} &  &  & &  & \multicolumn{1}{c|}{} &  & \textbf{\textcolor{green}{(+2.9)}} & \textbf{\textcolor{green}{(+10.2)}} &  \textbf{\textcolor{green}{(+1.1)}} \\ \hline

&& & \multicolumn{1}{l|}{} & \multicolumn{1}{c|}{}  & \multicolumn{1}{c|}{In-domain} & \multicolumn{12}{c|}{Out-of-domain} &  &  &  \\ \cline{6-18}
& \multicolumn{1}{l|}{\textbf{PT/MT=SDL}}& & & \multicolumn{1}{c|}{} & \multicolumn{1}{c|}{} & \multicolumn{12}{c|}{} &  &  &  \\ 

C0& \multicolumn{1}{l|}{ProtoNet \cite{Doersch2020}} && $\checkmark$& \multicolumn{1}{c|}{RN34} & \multicolumn{1}{c|}{53.7 ± 1.1} & 68.5 ± 1.3 & 58.0 ± 1.0 & 74.1 ± 0.9 & 68.8 ± 0.8 & 53.3 ± 1.1 & 40.7 ± 1.1 & 87.0 ± 0.7 & 58.1 ± 1.1 & 41.7 ± 1.1 &  &  & \multicolumn{1}{c|}{} & 53.7  & 61.1 & 61.1 & 60.4 \\
C1& \multicolumn{1}{l|}{CTX \cite{Doersch2020}} & & $\checkmark$&\multicolumn{1}{c|}{RN34} & \multicolumn{1}{c|}{62.8 ± 1.1} & 82.2 ± 1.0 & 79.5 ± 0.9 & 80.6 ± 0.9 & 75.6 ± 0.6 & 72.7 ± 0.8 & 51.6± 1.1 & 95.3 ± 0.4 & 82.7 ± 0.8 & 59.9 ± 1.0 &  &  & \multicolumn{1}{c|}{} & 62.8 & 75.6 & 75.6 & 74.3 \\
C2& \multicolumn{1}{l|}{TSA \cite{Li2021Cross-domainAdapters}} & &$\checkmark$& \multicolumn{1}{c|}{RN34} & \multicolumn{1}{c|}{63.7 ± 1.0} & 82.6 ± 1.1 & 80.1 ± 1.0 & 83.4 ± 0.8 & 79.6 ± 0.7 & 71.0 ± 0.8 & 51.4± 1.2 & 94.0 ± 0.5 & 81.7± 0.9 & 61.7 ± 0.9 & 94.6 ± 0.5 & 86.0 ± 0.6 & \multicolumn{1}{c|}{78.3 ± 0.8} & 63.7 &  76.2 & 78.7 & 77.5 \\
C3& \multicolumn{1}{l|}{DIPA} & $\checkmark$ & &\multicolumn{1}{c|}{ViT-s} & \multicolumn{1}{c|}{\textbf{71.4 ± 0.9}} & \textbf{84.3 ± 1.2} & \textbf{86.7 ± 1.0} & \textbf{88.2 ± 0.9} & \textbf{87.1 ± 0.6} & \textbf{74.6 ± 0.8} & \textbf{61.4 ± 1.2} & \textbf{97.4 ± 0.4} & \textbf{88.9 ± 1.0} & \textbf{65.2 ± 1.0} & \textbf{97.1 ± 0.5} & \textbf{88.5 ± 0.6} & \multicolumn{1}{c|}{\textbf{81.5 ± 0.8}} & \textbf{71.4}  & \textbf{81.5} & \textbf{83.4} & \textbf{82.5} \\ 

&& & \multicolumn{1}{l|}{} & \multicolumn{1}{c|}{} &  \multicolumn{1}{c|}{} &  &  & &  & & & \multicolumn{1}{c}{} &  &  & &  & \multicolumn{1}{c|}{} & \textbf{\textcolor{green}{(+7.7)}} & \textbf{\textcolor{green}{(+5.3)}} & \textbf{\textcolor{green}{(+4.7)}} & \textbf{\textcolor{green}{(+5.0)}} \\ 

& \multicolumn{1}{l|}{\textbf{PT=SDL-E, MT=SDL}} & && \multicolumn{1}{c|}{} & \multicolumn{1}{c|}{} & \multicolumn{12}{c|}{} &   &  &  \\ 
D0& \multicolumn{1}{l|}{PMF \cite{Hu2022PushingDifference}} & $\checkmark$ & $\checkmark$ &\multicolumn{1}{c|}{ViT-s} & \multicolumn{1}{c|}{74.7} & 80.7 & 76.8 & 85.0 & 86.6 & 71.3 & 54.8 & 94.6 & 88.3 & 62.6 &  &  & \multicolumn{1}{c|}{} & 74.7  & 77.9 & 77.9 & 77.5 \\
D1& \multicolumn{1}{l|}{DIPA} & $\checkmark$ && \multicolumn{1}{c|}{ViT-s} & \multicolumn{1}{c|}{\textbf{77.3 ± 0.7}} & \textbf{84.1 ± 1.2} & \textbf{87.1 ± 1.0} & \textbf{90.5 ± 0.7} & \textbf{87.3 ± 0.6} & \textbf{75.4 ± 0.8} & \textbf{60.9 ± 1.1} & \textbf{97.5 ± 0.4} & \textbf{91.7 ± 0.8} & \textbf{66.5 ± 0.9} & \textbf{97.2 ± 0.5} & \textbf{92.2 ± 0.5} & \multicolumn{1}{c|}{\textbf{84.5 ± 0.7}} & \textbf{77.3}  &\textbf{82.3} & \textbf{84.6} & \textbf{84.0} \\ 
& & &\multicolumn{1}{l|}{} & \multicolumn{1}{c|}{} & \multicolumn{1}{c|}{}& \multicolumn{1}{c}{} &  & &  & & & \multicolumn{1}{c}{} &  &  & &  & \multicolumn{1}{c|}{} & \textbf{\textcolor{green}{(+2.6)}} & \textbf{\textcolor{green}{(+4.4)}} & \textbf{\textcolor{green}{(+6.7)}} & \textbf{\textcolor{green}{(+6.5)}} \\ \hline

\end{tabular}%
}
\caption{Comparison of state-of-the-art methods on Meta-Dataset using MDL, SDL and SDL-E settings where PT: Pre-training, MT: Meta-Training, RN: ResNet, ViT-s: ViT-small, com.: common, Avg.: Average, SS PT: indicates self-supervised pre-training and Sup. MT: indicates supervised meta-training. Mean accuracy and 95\% confidence interval are reported, where available. }
\label{tab:MDL_results}
\end{table*}

From here onwards, we refer to our method as \textit{DIPA}, \textbf{DI}scriminative-sample-guided and \textbf{P}arameter-efficient feature \textbf{A}daptation. We evaluate DIPA with the feature extractors pre-trained under the MDL, SDL and SDL-E settings on Meta-Dataset and compare it to existing state-of-the-art methods in Table \ref{tab:MDL_results}. To facilitate fair comparisons, given the varying meta (or pre)-training strategies and backbone architectures in prior research, we include these details alongside the results. The table is organized into two sections: one for in-domain (seen) and another for out-of-domain (unseen) dataset accuracies, including their overall averages. Additionally, we include a column for the average accuracy across commonly unseen domains, as some prior works do not report results for all the unseen domains considered in our work.

\textbf{Multi-domain feature extractor}. In Table \ref{tab:MDL_results} (rows A0-A6), we benchmark against state-of-the-art methods that meta (or pre)-train solely under the MDL setting. Notably, DIPA outperforms eTT \cite{Xu2023ExploringTransformers}, which employs self-supervised pre-training, across all common seen and unseen domains (10 out of 10) and is up to $32.5\times$ more parameter-efficient (refer Table \ref{tab:param_counts}, $d_t=7$ and $9$). When compared to methods using supervised meta-training for their feature extractor \(f_\theta\) (such as Simple CNAPS \cite{Bateni2022BeyondLearning}, URT \cite{Liu2020AClassification}, FLUTE \cite{Triantafillou2021LearningGeneralization}, TSA \cite{Li2021Cross-domainAdapters}, and SSA \cite{SreenivasSimilarLearning}), DIPA shows superior performance in most domains, particularly outperforming them in 4 out of 5 unseen domains. Improving performance in unseen domains is a significant challenge due to the vast difference between seen and unseen domains and the limited availability of labelled samples for new tasks. DIPA addresses this by employing lightweight linear transformations for feature adaptation, together with a discriminative sample-guided loss function. While TSA+SSA \cite{SreenivasSimilarLearning} also shows competitive results using MixStyle-like augmentation strategies \cite{zhou2021domain}, DIPA achieves even better performance in most unseen domains without such augmentations\footnote{Incorporating similar augmentation strategies as in \cite{SreenivasSimilarLearning} into DIPA could lead to further improvements, a prospect we reserve for future exploration.} while being up to $24.7\times$ more parameter-efficient (see Table \ref{tab:param_counts}). Specifically, DIPA significantly surpasses TSA+SSA\cite{SreenivasSimilarLearning} in Traffic Sign (+6.4), MS-COCO (+6.7), CIFAR-10 (+4.5), and CIFAR-100 (+10.4).

\textbf{Multi-domain feature extractor with additional pre-training}. In Table \ref{tab:MDL_results} (rows B0-B2), we compare DIPA with PMF \cite{Hu2022PushingDifference} and ATTNSCALE \cite{Basu2023StrongBaseline}, which evaluate their models under the MDL setting and use SDL-E for self-supervised pre-training. Note that, unlike DIPA and ATTNSCALE, PMF also employs MDL for meta-training. DIPA outperforms ATTNSCALE, achieving a 2.9\% higher performance in unseen domains and 2.7\% in seen domains, and is up to $4.3\times$ more parameter-efficient (see Table \ref{tab:param_counts}). Despite PMF incorporating an additional meta-training stage and task-specific learning rate selection\cite{Hu2022PushingDifference}, DIPA exceeds PMF's unseen domain performance by 3.3\%. This enhanced performance is achieved with significant computational efficiency, as DIPA requires only two stages: pre-training and fine-tuning, thus eliminating the need for additional meta-training or learning rate selection. Furthermore, DIPA's fine-tuning process is remarkably efficient, using only up to 0.08\% of the model parameters, a stark contrast to PMF’s 100\% parameter utilization (see Table \ref{tab:param_counts}).


\textbf{Single-domain feature extractor}. We also evaluate our method using a feature extractor that is trained solely on the ImageNet domain \cite{russakovsky2015imagenet}, referred to as the single-domain learning (SDL) setting. The SDL setting poses greater challenges compared to multi-domain scenarios, given that the model is exposed to only one training domain but evaluated across multiple, including the test split of ImageNet and additional diverse domains. In Table \ref{tab:MDL_results} (rows C0-C3), we benchmark our method against existing methods (ProtoNet \cite{snell2017prototypical}, CTX \cite{doersch2020crosstransformers}, and TSA \cite{Li2021Cross-domainAdapters}) with published results under this setting. Our method surpasses these methods across all 13 domains. It is important to acknowledge that our model utilizes self-supervised pre-training together with a backbone architecture that is substantially different from those compared here, which may affect the fairness of this comparison. Nonetheless, our method still demonstrates significant performance gains, with improvements of 7.7\%, 4.7\%, and 5\% for seen, unseen (all), and all domains, respectively.

\textbf{Single-domain feature extractor with additional pre-training}. In Table \ref{tab:MDL_results} (rows D0-D1), we present a comparison of DIPA with the latest state-of-the-art method, PMF \cite{Hu2022PushingDifference}. Here, PMF employs SDL-E for pre-training and an additional SDL stage for meta-training, a step not used in DIPA. Despite the absence of this meta-training stage, DIPA demonstrates superior performance across all 13 domains. Specifically, it shows a 2.6\% improvement in seen domains, 4.4\% in commonly unseen domains, and 6.5\% across all domains. This performance is achieved without requiring additional computational steps or the extensive fine-tuning of the entire backbone, as is necessary for PMF (see details in Table \ref{tab:param_counts}).

Overall, our results demonstrate that DIPA significantly outperforms state-of-the-art methods in cross-domain scenarios within both seen and unseen domains while being more parameter efficient.

\section{Additional Analysis}
\label{sec:additional_analysis}

In this section, we conduct ablation studies to evaluate each step in our method and validate their impact on improving the few-shot classification performance.

\subsection{Impact of varying the fine-tuning and query classification approaches}
\label{sec:evaluatingncc_vs_variance}

\begin{table*}[!ht]
\centering
\resizebox{\textwidth}{!}{%
\begin{tabular}{c|c|c|cccccccc|ccccc|ccc}
\hline
& Fine-tune & Inference & ImageNet             & Omniglot             & Aircraft             & Birds                & Textures             & QuickDraw           & Fungi                & VGGFlower           & TrafficSign         & MSCOCO              & MNIST                & CIFAR10             & CIFAR100            & Avg. & Avg.  & Avg.     \\ 

& stage & stage &               &              &              &                 &              &            &                 &            &          &               &                 &              &             & Seen & Unseen & All    \\ \hline
A0 & $NCC_{mean}$       & $NCC_{mean}$                & 68.0 ± 1.0          & 79.0 ± 1.4          & 81.8 ± 1.1          & 88.2 ± 0.9          & 87.3 ± 0.6          & 73.2 ± 0.8          & 63.4 ± 1.0          & 97.2 ± 0.4          & 89.0 ± 1.0          & 63.3 ± 1.0          & 95.2 ± 0.6          & 87.2 ± 0.7          & 78.8 ± 0.9         & 79.8	& 82.7 &	80.9 \\
A1 & $l_{A_\phi}$        & $NCC_{A_\phi}$                & 65.5 ± 1.0          & 83.4 ± 1.1          & 81.8 ± 1.0          & 88.6 ± 0.8          & 87.0 ± 0.6          & 68.4 ± 0.9          & 57.0 ± 1.2          & 97.5 ± 0.3          & 86.7 ± 1.1          & 52.4 ± 1.2          & 96.8 ± 0.5          & 82.8 ± 0.8          & 76.3 ± 0.9          & 78.7	& 79.0	& 78.8  \\
A2 & $l_{A_\phi}$        & $NCC_{mean}$                & \textbf{70.9 ± 1.0} & \textbf{84.7 ± 1.1} & \textbf{86.3 ± 1.0}          & \textbf{90.8 ± 0.8} & \textbf{88.6 ± 0.5} & \textbf{75.3 ± 0.8}          & \textbf{66.6 ± 1.1} & \textbf{97.9 ± 0.3} & \textbf{91.3 ± 1.0} & \textbf{64.8 ± 1.0}          & \textbf{96.9 ± 0.5} & \textbf{87.4 ± 0.6}          & \textbf{81.2 ± 0.8}          & \textbf{82.6}	& \textbf{84.3}	& \textbf{83.3} \\   \hline

\end{tabular}%
}
\caption{Comparing the impact of using the traditional NCC ($NCC_{mean}$) vs. $l_{A_\phi}$ loss for fine-tuning and $NCC_{mean}$ vs anchor-based NCC ($NCC_{A_\phi}$) for inference. The results are reported for the MDL setting where the first eight datasets are seen during training and the last five datasets are unseen and used for testing only. Mean accuracy and 95\% confidence interval are reported. }
\label{tab:classifier_comparison}
\end{table*}

First, we compare our fine-tuning strategy \(l_{A_\phi}\) against the conventional mean-embedding-based NCC (\(NCC_{mean}\)) in Table \ref{tab:classifier_comparison} (rows A0 and A2). Here, employing $l_{A_\phi}$ for fine-tuning significantly improves classification performance, yielding up to 2.8\% improvement in seen domains and 1.6\% in unseen domains compared to using \(NCC_{mean}\). Moreover, in terms of cluster formation, on the ImageNet dataset, $l_{A_\phi}$ consistently outperformed \(NCC_{mean}\) in terms of inter-cluster distance, intra-cluster distance, and silhouette index by 1\%,
6\% and 1\%, respectively.

Next, we compare our query classification strategy -  \(NCC_{mean}\) - with another alternative. As mentioned above, during fine-tuning, DIPA learns a set of anchors $A_\phi$ that provide strong supervision to form well-separated and compact clusters (refer Fig. \ref{fig:cluster_vis_training}b). This provides an opportunity to use the fine-tuned anchors in lieu of the mean class centroids used in \(NCC_{mean}\) for query classification. We denote this variant as \(NCC_{A_\phi}\). However, while anchors $A_\phi$ provide strong supervisory signals for cluster formations, we observe that the anchors are placed with a small offset from the mean centroid in some cases (refer Supplementary Fig. \ref{appendix_fig:clusters_3}), suggesting that the mean class centroid which encapsulates the summary of the compact cluster might be a more representative class descriptor during inference. This hypothesis is also supported by our results, where combining \(l_{A_\phi}\) with \(NCC_{mean}\) yielded an optimal performance increase of 3.9\% and 5.3\% on seen and unseen domains compared to \(l_{A_\phi}\)+\(NCC_{A_\phi}\). Therefore, we employ \(A_\phi\) anchors for cluster formation during fine-tuning and thereafter, use the well-defined clusters to obtain mean class centroids for query classification during inference.

\subsection{Impact of varying the number of tuned layers}
\label{sec:tuned_layer}

\begin{figure}[ht]
    \centering
    \includegraphics[width=\columnwidth]{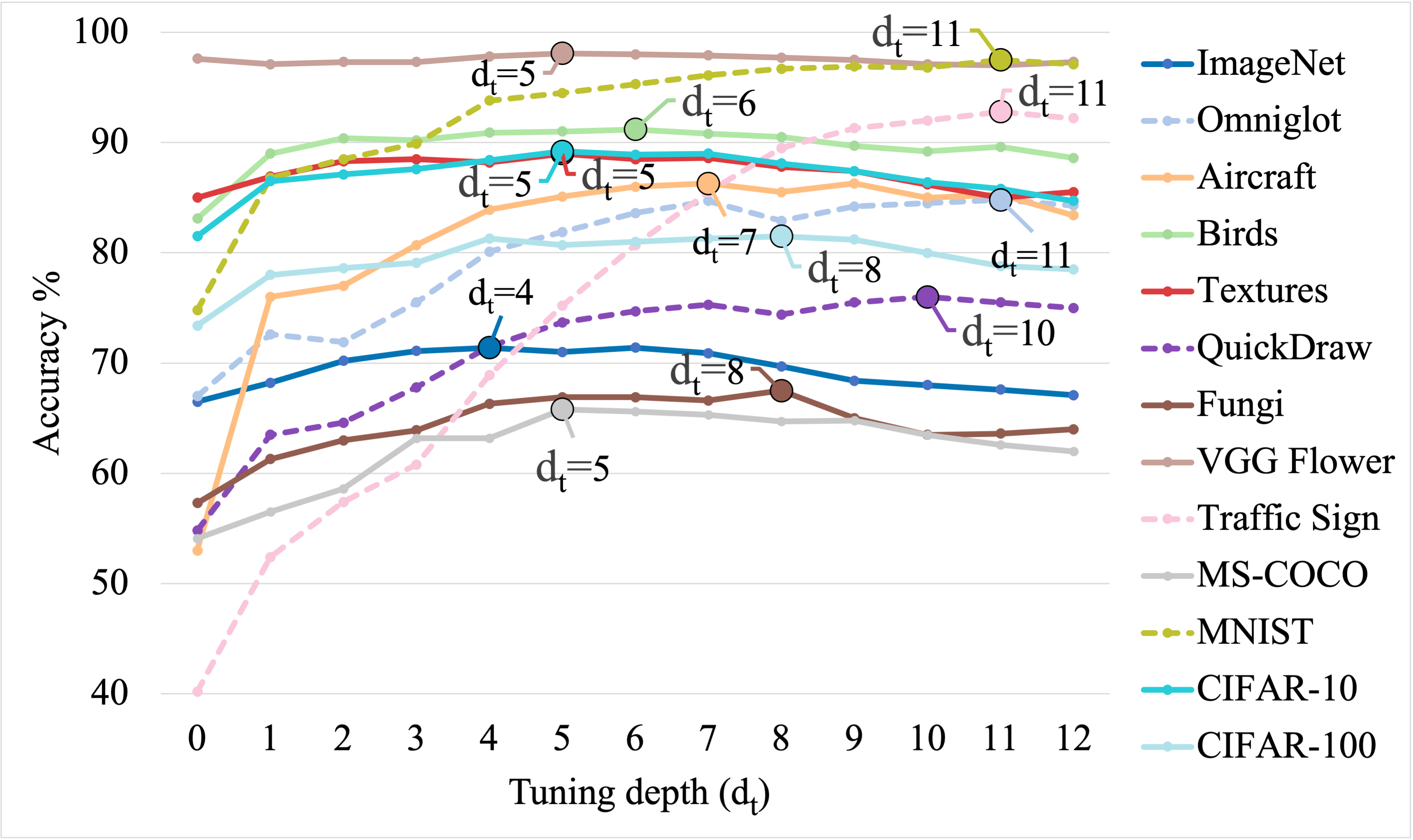}
    \caption{Variation of accuracies for each dataset in Meta-Dataset as $d_t$ varies in the MDL setting. Average results are reported while more detailed results can be found in Supplementary Table \ref{appendix_tab:tuning_depth_datasets}. The dotted lines represent the relatively more challenging datasets. For each dataset, the value of $d_t$ that reports the highest accuracy is annotated with a dot.}
    \label{fig:tunable_depth_domain_specific}
\end{figure}

We study the impact of varying the number of tuned layers in $f_\theta$. Fig. \ref{fig:tunable_depth_domain_specific} illustrates the average accuracy variation for each dataset in Meta-Dataset as we vary the number of adapted layers $d_t$, where $0 \leq d_t \leq L$ and $L=12$ for ViT-small. From the results, we observe the following: 1) A universal value of $d_t$ that provides optimal results for all domains is not available, 2) A general trend is present where domains that are relatively less challenging such as VGG Flower, CIFAR-10 etc., report higher accuracies for smaller values of $d_t$. In contrast, the relatively more challenging domains, such as Quickdraw, and Traffic Sign, report higher accuracies for larger values of $d_t$, 3) Most domains report higher accuracies with $d_t<12$. i.e., our approach does not need task-specific parameters attached to each layer of the model (as in TSA \cite{doersch2020crosstransformers}, and eTT \cite{Xu2023ExploringTransformers}) nor fine-tuning the entire backbone (as in PMF \cite{Hu2022PushingDifference}). This facilitates an even further reduction in the count of trainable parameters. For example, as shown in Table \ref{tab:param_counts}, by reducing $d_t$ from 12 to 7, we achieve a reduction in the parameter count by 0.2\%. 

\begin{figure}[ht]
    \centering
    \includegraphics[width=\columnwidth]{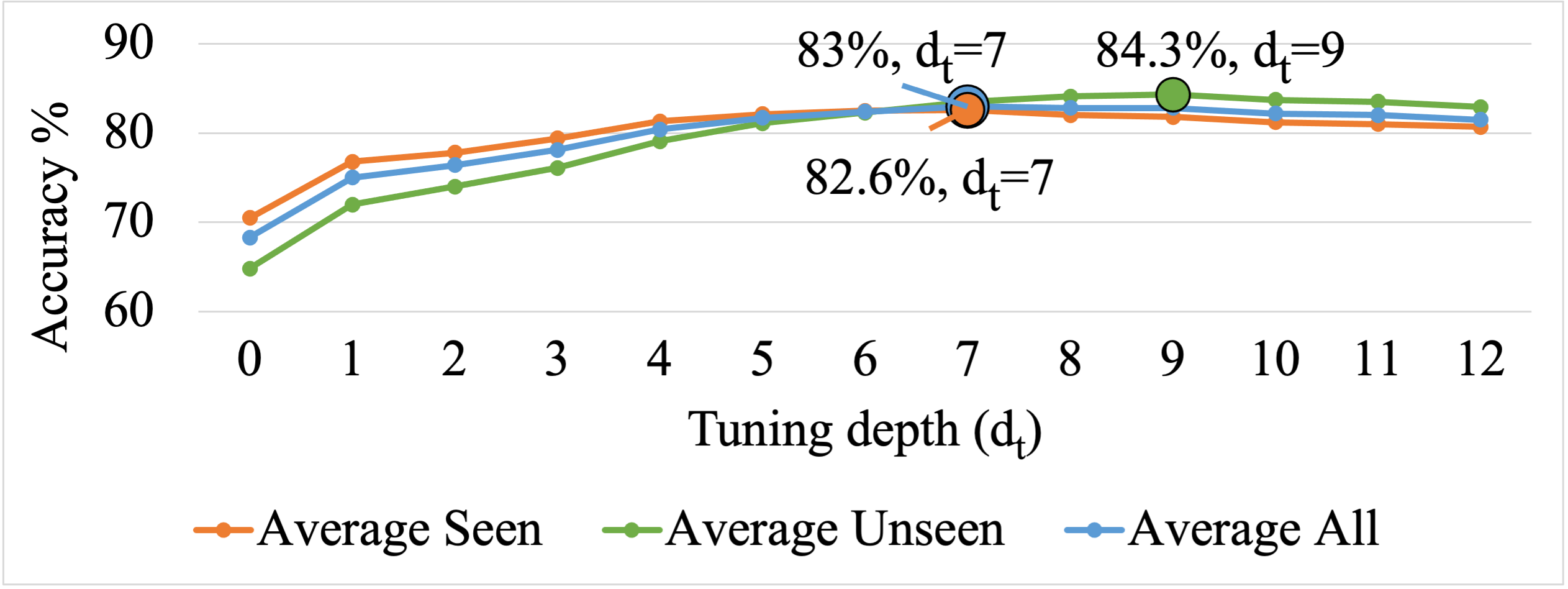}
    \caption{Variation of average seen, unseen and total average accuracies across all domains in Meta-Dataset as the number of tuned layers $d_t$ varies in the MDL setting. Average results are reported and more detailed results can be found in Supplementary Table \ref{appendix_tab:tuning_depth_datasets}. The value of $d_t$ that reports the highest average for each line is annotated with a dot.}
    \label{fig:tunable_depth_average_MDL}
\end{figure}

Fig. \ref{fig:tunable_depth_domain_specific} demonstrates that customizing the depth \( d_t \), on a per-domain basis, yields the best performance. This customization, however, adds extra user-defined parameters to our model. To simplify, we streamline our approach to use just two \( d_t \) values: \( d_t=7 \) for seen domains and \( d_t=9 \) for unseen domains, based on peak average performance as illustrated in Fig. \ref{fig:tunable_depth_average_MDL}. Moreover, opting for a higher \( d_t \) value for unseen domains aligns with the intuition that a more extensive adaptation, via tuning additional layers, is typically beneficial when dealing with unseen domains.

\subsection{Impact of varying the feature fusion depth}

We compare the performance variation for few-shot classification by varying the feature fusion depth $d_f$ and report the average results in Table \ref{tab:fusion-depth}. First, we note that even without feature fusion (when $d_f=1$), our method outperforms existing methods by a significant margin. Next, we observe that as we increase $d_f$ in DIPA, the average accuracies generally show an increasing trend up to $d_f=4, 6$, validating our choice of fusing features from multiple layers. However, the performance shows a gradual declining trend afterwards. One possible reason is that combining a higher degree of shallower layers (containing more generic patterns) will cause the model to lose focus on important domain-specific features in the deeper layers. This is also reflected in the results where $d_f=8, 12$ report even lower accuracies than $d_f=1, 2$.

\begin{table}[!htbp]
\centering
\resizebox{.7\columnwidth}{!}{
\begin{tabular}{c|c|c|ccc}
\hline
Method & $d_f$ & $dim$ & Avg. Seen  & Avg. Unseen & Avg. all   \\ \hline
TSA + SSA \cite{SreenivasSimilarLearning} & 1 & 512 & 80.8 & 79.0 & 80.1 \\
eTT \cite{Xu2023ExploringTransformers} & 1 & 384 & 79.6 & 73.7 & 78.3 \\ 
\hline
 & 1  & 384 & 82.3          & 83.7           & 82.9          \\
 & 2  & 768 & 82.4          & 83.7           & 82.9          \\
DIPA & 4  & 1536 & \textbf{82.6} & \textbf{84.3}  & \textbf{83.3} \\
 & 6  & 2304 & \textbf{82.6}          & 83.9           & 83.1          \\ 
 & 8  & 3072 & 82.0          & 83.5           & 82.6          \\
 & 12 & 4608 & 81.5          & 83.2           & 82.1         \\ \hline
\end{tabular}%
}
\caption{Variation of average (Avg.) accuracies for seen, unseen and all domains as the feature fusion depth $d_f$ and its corresponding feature representation's dimension ($dim$) under the MDL setting. Mean results are reported and more detailed results can be found in Supplementary Table \ref{appendix_tab:fusion-depth}. }
\label{tab:fusion-depth}
\end{table}

\subsection{Impact of Pre-training}

In Table \ref{tab:pre-train-results}, we evaluate the impact of using MIM for pre-training, as opposed to the more commonly used DINO pre-training in existing methods \cite{Hu2022PushingDifference, Xu2023ExploringTransformers, Basu2023StrongBaseline}. For our experiments, we leveraged the SDL-E setting, utilizing the readily available DINO checkpoint \cite{Caron2021EmergingTransformers}. First, we compare the DIPA model with DINO pre-training to the DINO-based PMF \cite{Hu2022PushingDifference} model, which represents the state-of-the-art in the SDL-E setting. As shown in the first two rows of Table \ref{tab:pre-train-results}, DIPA improves performance by 5.8\% compared to PMF, demonstrating its effectiveness even without our preferred sel-supervised pre-training task. Next, we substitute the DINO-based feature extractor with a MIM-based one (refer Table \ref{tab:pre-train-results} row 3). This change leads to a further increase in accuracy up to 6.5\% over PMF, validating the effectiveness of our selected pre-training algorithm.

\begin{table}[!ht]
\centering
\resizebox{\columnwidth}{!}{%
\begin{tabular}{cc|l|l|l|l}
\hline
Method & Pre-Train  & Avg. Seen & Avg. Unseen (com.) & Avg. Unseen (all) & Avg. All  \\ \hline
PMF  \cite{Hu2022PushingDifference} & DINO & 74.7 & 77.9 & 77.9 & 77.5 \\
DIPA & DINO  & 75.9 \textcolor{green}{(+1.2)} & 82.0 \textcolor{green}{(+4.1)} & 83.9 \textcolor{green}{(+6.0)}  &  83.3 \textcolor{green}{(+5.8)}    \\ 
DIPA & MIM &  \textbf{77.3} \textcolor{green}{(+2.6)} & \textbf{82.3} \textcolor{green}{(+4.4)} & \textbf{84.6} \textcolor{green}{(+6.7)}  & \textbf{84.0} \textcolor{green}{(+6.5)}  \\ \hline          
\end{tabular}%
}
\caption{The impact of varying the pre-training algorithms in SDL-E setting. Average (Avg.) results are reported and more detailed results can be found in Supplementary Table \ref{appendix_tab:pre_training}.}
\label{tab:pre-train-results}
\end{table}

%% file: sec/6_conclusions.tex
\section{Conclusions and Limitations}

In this study, we explore the efficient adaptation of neural networks for few-shot classification. Our approach leverages extremely lightweight linear transformations, optimized by a discriminative sample-aware loss function, to learn new  classes and domains with a limited number of labelled samples. This method achieves state-of-the-art performance on the challenging Meta-Dataset benchmark while ensuring parameter efficiency.

Our method is not without limitations. The current approach applies a fixed linear transformation to every layer of the pre-trained model. Future improvements could enable these transformations to be defined flexibly, layer by layer, to suit the specific requirements of the target task. Moreover, instead of restricting tuning depth to only two values for seen and unseen datasets, which may lead to suboptimal outcomes, future research could explore defining optimal tuning depths customized for each dataset and task.


\section{Acknowledgements}
R.P. acknowledges the Melbourne Graduate Research scholarship and GCI Top-up scholarship. S.H. acknowledges ARC grant DP210101135. The authors thank Hai-Hang Wu, Tamasha Malepathirana, Nisal Ranasinghe and Maneesha Perera for proofreading.

%% file: sec/x_supplementary.tex
\clearpage
\setcounter{page}{1}
\maketitlesupplementary

\section{Datasets}
\label{appendix:datasets}
\textbf{Meta-Dataset} \cite{Triantafillou2020} is a few-shot classification benchmark that initially consists of ten datasets: ILSVRC\_2012 (ImageNet) \cite{russakovsky2015imagenet}, Omniglot \cite{doi:10.1126/science.aab3050}, FGVC-Aircraft (Aircraft) \cite{maji2013fine}, CUB-200-2011 (Birds) \cite{wah2011caltech}, Describable Textures (Textures) \cite{cimpoi2014describing}, QuickDraw \cite{jongejan2016quick}, FGVCx Fungi (Fungi) \cite{brigit2018fgvcx}, VGG Flower \cite{nilsback2008automated}, Traffic Signs \cite{houben2013detection} and MS-COCO \cite{lin2014microsoft}. It further expands with the addition of MNIST \cite{lecun1998gradient}, CIFAR-10 \cite{krizhevsky2009learning} and CIFAR-100 \cite{krizhevsky2009learning}. Each dataset is further divided into train, validation and test sets with disjoint classes. We follow the standard training protocols proposed by \cite{Triantafillou2020} and consider both ``Training on all datasets" (MDL: multi-domain learning) and ``Training on ImageNet-Train only" (SDL: single-domain learning) settings. For the former, we follow the standard procedure and use the training set of the first eight datasets for pre-training. During evaluation, the test set of the eight datasets are used for evaluating the generalization ability in the seen domains while the remaining five datasets are used to evaluate the cross-domain generalization ability. In the ``Training on ImageNet-Train only" setting, we follow the standard procedure and only use the train set of ImageNet for pre-training. The evaluation is performed on the test set of ImageNet as the seen domain while the rest 12 datasets are considered unseen domains. Additionally, to compare our method with more recent state-of-the-art \cite{Hu2022PushingDifference}, we also use a pre-trained model on the full ImageNet dataset for ``Training on ImageNet-Full" (SDL-E: single-domain learning-extra data) setting, where the evaluation is performed similarly to the ``Training on ImageNet-Train only" setting.

We also report additional results for the following datasets in the Appendix. 

\textbf{miniImageNet} \cite{Vinyals2016} contains 100 classes from ImageNet-1k, set into 64 training, 16 validation and 20 testing classes. 

\textbf{CIFAR-FS} \cite{cifarfs2019} is created by dividing the original CIFAR-100 into 64 training, 16 validation and 20 testing classes.

\section{Implementation details}
\label{appendix:Implementation}

\subsection{Pre-trianing using Masked Image Modelling}
\label{appendix:Pretrianing}

We employ Masked Image Modelling (MIM) to pre-train the feature extractor and follow the hyper-parameters and data augmentations recommended in  \cite{Zhou2021IBOT:Tokenizer}. The teacher patch temperature was set to 0.04, in contrast to the default value of 0.07 after observing that a lower temperature leads to more consistent and stable training losses.\footnote{https://github.com/bytedance/ibot/issues/19}. 

\textbf{MDL:} We employ the train sets of the eight in-domain datasets (ImageNet, Omniglot, Aircraft, Birds, Textures, and VGG Flower) considered under the MDL setting for pre-training $f_\theta$.

\textbf{SDL:} We employ the train set of the ImageNet dataset for pre-training $f_\theta$.

\textbf{SDL-E:} The entire ImageNet dataset is utilized to train the feature extractor \footnote{https://github.com/bytedance/ibot\#pre-trained-models} $f_\theta$ \cite{Hu2022PushingDifference}. 

\subsection{Pre-trianing on DINO}
 \textbf{SDL-E:} To compare with the MIM pre-training on SDL-E, we utilize the pre-trained checkpoint weights provided by DINO \cite{Caron2021EmergingTransformers} after training on the entire ImageNet dataset.

\subsection{Hyperparameters}
\label{appendix:Hyperparameters}

\subsubsection{Task-specific parameter initialization}
\label{appendix:Taskspecificparameter}

For results reported in the main text, we choose constant initialization of task-specific parameters $\gamma$ (scale) and $\beta$ (shift) as one and zero, respectively. However, one could also employ a normalized initialization, where the mean values of $\gamma$ and $\beta$ are one and zero \cite{Zhou2021IBOT:Tokenizer}. Therefore, we report results for normalized initialization and constant initialization in Supplementary Table \ref{appendix_tab:intializations} columns 1 and 2, respectively. Notably, we obtain better results for the constant initialization in comparison to the normalized initialization.

\begin{table}[!htbp]
\centering
\resizebox{\columnwidth}{!}{%
\begin{tabular}{l|cccc}
\hline

$\gamma, \beta$ (constant) & &   $\checkmark$ & $\checkmark$ & $\checkmark$ \\ 
$\gamma, \beta$ (normal)  & $\checkmark$ & & & \\ 
NAdam & $\checkmark$ & $\checkmark$ & & $\checkmark$ \\ 
AdamW & &  & $\checkmark$ & \\ 
${A_\phi}$ (random) & $\checkmark$ & $\checkmark$ & $\checkmark$ & \\ 
${A_\phi}$ (custom) & & & & $\checkmark$ \\ \hline

ImageNet & 69.22 ± 0.94 & \textbf{70.86 ± 0.95} & 68.21 ± 0.96 & 70.25 ± 0.98 \\
Omniglot & 83.55 ± 1.17 & \textbf{84.68 ± 1.10} & 82.79 ± 1.18 & 84.55 ± 1.15 \\
Aircraft & 85.91 ± 1.06 & 86.33 ± 0.95 & \textbf{86.55 ± 1.00} & 85.05 ± 1.06 \\
Birds & 90.31 ± 0.80 & \textbf{90.75 ± 0.75} & 88.49 ± 0.88 & 89.70 ± 0.88 \\
Textures & 87.66 ± 0.66 & 88.60 ± 0.51 & 87.15 ± 0.62 & \textbf{88.61 ± 0.56} \\
Quick Draw & 74.27 ± 0.82 & \textbf{75.29 ± 0.77} & 72.81 ± 0.83 & 75.10 ± 0.77 \\
Fungi & 66.07 ± 1.05 & \textbf{66.64 ± 1.05} & 64.30 ± 1.05 & 65.54 ± 1.07 \\
VGG Flower & 97.71 ± 0.32 & \textbf{97.88 ± 0.30} & 97.24 ± 0.38 & 97.63 ± 0.32 \\ \hline
Traffic Sign & 89.84 ± 1.19 & \textbf{91.29 ± 0.96} & 87.29 ± 1.13 & 89.80 ± 0.97 \\
MS-COCO & 62.34 ± 1.04 & \textbf{64.78 ± 0.95} & 57.84 ± 1.07 & 64.67 ± 1.01 \\
MNIST & 96.64 ± 0.49 & \textbf{96.87 ± 0.53} & 96.14 ± 0.60 & 96.82 ± 0.50 \\
CIFAR-10 & 84.56 ± 0.85 & 87.40 ± 0.64 & 79.72 ± 1.03 & \textbf{87.81 ± 0.66} \\
CIFAR-100 & 79.38 ± 0.94 & \textbf{81.24 ± 0.78} & 75.29 ± 0.94 & 80.28 ± 0.83 \\ \hline
Average Seen & 79.6	&	\textbf{82.6}	&	80.9	&	82.1 \\
Average Unseen & 82.6	&	\textbf{84.3}	&	79.3	&	82.7 \\
Average All & 80.8	&	\textbf{83.3}	&	80.3	&	82.3 \\ \hline
\end{tabular}%
}
\caption{Comparision of varying the task-specific parameter initialization (constant vs normal), Optimizers (NAdam vs AdamW) and ${A_\phi}$ anchor initialization (random vs custom) in the MDL setting.}
\label{appendix_tab:intializations}
\end{table}

\subsubsection{AdamW vs NAdam}
\label{appendix:optimizers}
Recently, AdamW has gained popularity as a preferred choice for fine-tuning large models such as ViTs \cite{Xu2023ExploringTransformers}. Nevertheless, our results, as detailed in Supplementary Table \ref{appendix_tab:intializations}, columns 2 and 3, demonstrate that NAdam yields superior performance in the context of cross-domain few-shot classification.


\subsubsection{Anchor initialization}
\label{appendix:varianceinit}
While ${A_\phi}$ anchors in DIPA are randomly initialized for each task, one can argue that using the mean of the support embedding vectors can be a favourable anchor initialization point. Consequently, we report the results for random vs custom initialization of anchors in Supplementary Table \ref{appendix_tab:intializations}, columns 2 and 4. Here, the anchors are randomly initialized for ${A_\phi}$ (random) while the mean of class embedding vectors initializes the anchors for ${A_\phi}$ (custom). Notably, using unadapted feature embeddings for anchor initialization may hinder fine-tuning due to priors imposed by unadapted features. In contrast, using random initialization, together with a substantial learning rate may offer better adaptability for anchors during fine-tuning without being influenced by irrelevant priors. This is also reflected in the results reported in Supplementary Table \ref{appendix_tab:intializations}, columns 2 and 4, where random initialization outperforms custom initialization, confirming our selection in the DIPA framework. 

\subsubsection{Number of fine-tuning iterations}
\label{appendix:finetuneepoch}
We experimentally determined the number of fine-tuning iterations. We report the results for four such scenarios in Supplementary Table \ref{appendix_tab:HYPERPARAM_EPOCH}. As reported in the results, $l_{A_\phi}$ with 80 iterations provides the highest accuracy. Therefore, in our framework, we use 80 as the number of iterations for fine-tuning. 

\begin{table}[!htbp]
\centering
\resizebox{\columnwidth}{!}{
\begin{tabular}{lcccc} \hline
Fine Tune                     & $\#$ Iterations           & \multicolumn{1}{l}{Avg. Seen} & Avg. Unseen  & Avg. All        \\ \hline
DIPA & 40 & 81.5 & 82.4 & 81.8  \\
\textbf{DIPA} & \textbf{80} & \textbf{82.6} & \textbf{84.3} & \textbf{83.3}  \\ \hline
\end{tabular}%
}
\caption{Comparing the average (Avg.) performance variation as the number of epochs varies for seen, unseen and all domains in the MDL setting.}
\label{appendix_tab:HYPERPARAM_EPOCH}
\end{table}

\section{Addtional Results for Meta-Dataset}

\subsection{Feature Space Visualizations: Before and After Fine-Tuning}
\label{appendix:FeatureVisualizations}

By using UMAP visualizations, we identify the impact of fine-tuning the feature space using $l_{A_\phi}$ in Supplementary Fig. \ref{appendix_fig:clusters_1} and \ref{appendix_fig:clusters_2}. Here, the left columns illustrate that semantic clusters have already emerged using the pre-trained MIM features, although overlapped/dispersed in some instances. Thereafter, as illustrated in the right columns, $l_{A_\phi}$ uses the strong initialization provided by MIM and further refines the feature space to form better-separated clusters that show high inter-class variance and low intra-class variance. 

\begin{figure*}[]
  \centering
   \includegraphics[width=1\textwidth]{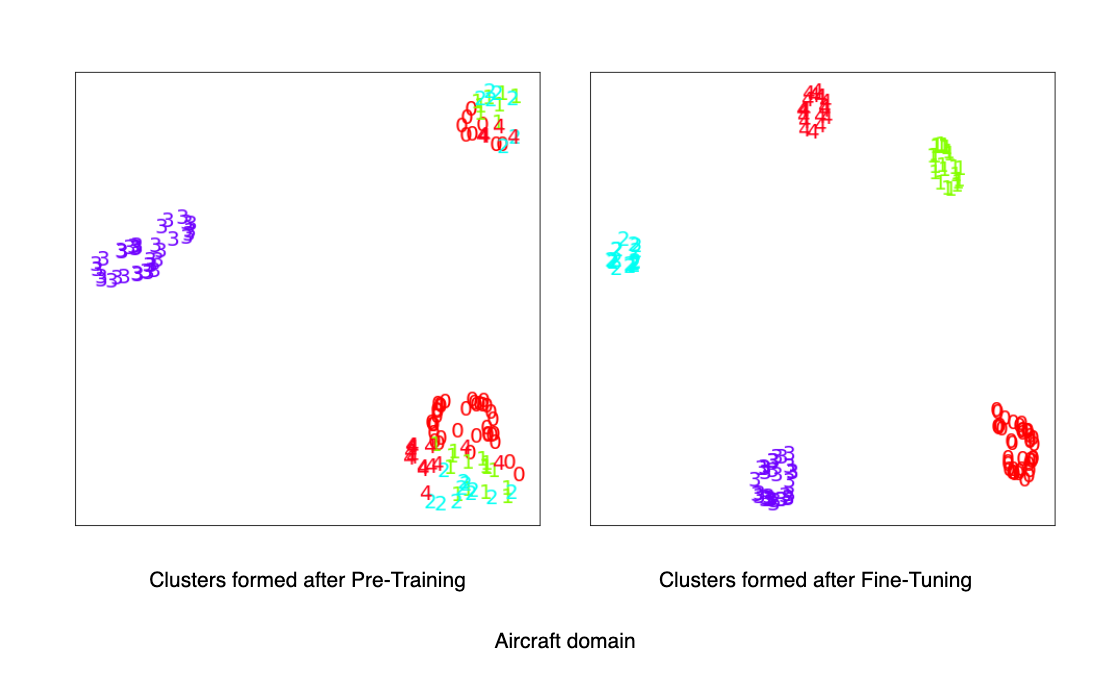}
   \caption{UMAP visualization of clusters formed in the feature space for Aircraft domain in MDL setting. The clusters formed before and after fine-tuning with DIPA, are illustrated in the first and second columns, respectively.}
   \label{appendix_fig:clusters_1}
\end{figure*}

\begin{figure*}[]
  \centering
   \includegraphics[width=1\textwidth]{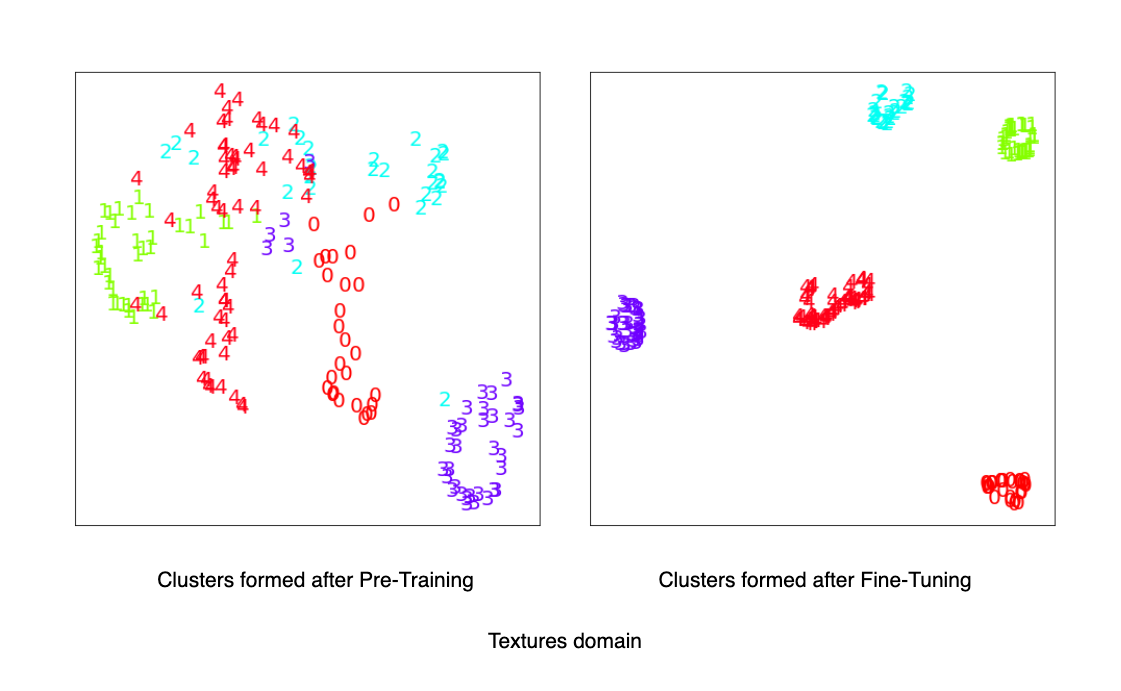}
   \caption{UMAP visualization of clusters formed in the feature space for Textures domain in MDL setting. The clusters formed before and after fine-tuning with DIPA are illustrated in the first and second columns, respectively.}
   \label{appendix_fig:clusters_2}
\end{figure*}

\subsection{Prototype Visualizations}

\begin{figure}[]
  \centering
   \includegraphics[width=1\columnwidth]{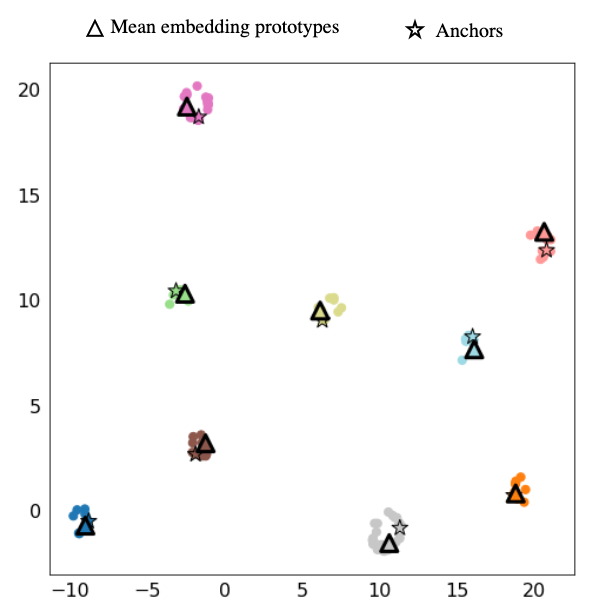}
   \caption{UMAP visualization of clusters formed in the feature space after fine-tuning with mean embedding-based prototypes and anchors $A_\phi$.  }
   \label{appendix_fig:clusters_3}
\end{figure}

The placement of anchors $A_\phi$ and mean embedding-based prototypes after fine-tuning is visualized in Fig. \ref{appendix_fig:clusters_3}. As discussed in Section \ref{sec:evaluatingncc_vs_variance}, while $A_\phi$ provides strong supervision for cluster formation during fine-tuning, after fine-tuning, we observe that they are placed with a small offset from the mean representation (mean embedding prototype) of the clusters. 

\subsection{Impact of tuned depth}
\label{appendix:tuneddepth}

Supplementary Table \ref{appendix_tab:tuning_depth_datasets} reports the variation of accuracies as the number of tuned layers $d_t$ vary on the MDL setting for Meta-Dataset. A summary of Supplementary Table \ref{appendix_tab:tuning_depth_datasets} is shown in the main text's Fig. \ref{fig:tunable_depth_domain_specific} and \ref{fig:tunable_depth_average_MDL}.

\begin{table*}[]
\centering
\resizebox{\textwidth}{!}{%
\begin{tabular}{l|ccccccccccccc}
\hline
 \multicolumn{1}{c|}{$d$} & 0 & 1 & 2 & 3 & 4 & 5 & 6 & 7 & 8 & 9 & 10 & 11 & 12 \\ \hline
 ImageNet & 66.51 ± 1.02 & 68.22 ± 0.95 & 70.24 ± 1.01 & 71.11 ± 0.95 & 71.37 ± 0.94 & 71.00 ± 0.92 & 71.36 ± 0.91 & 70.86 ± 0.95 & 69.71 ± 0.94 & 68.39 ± 0.95 & 68.05 ± 0.92 & 67.57 ± 0.95 & 67.13 ± 0.93 \\
 Omniglot & 67.04 ± 1.23 & 72.63 ± 1.29 & 71.87 ± 1.34 & 75.52 ± 1.25 & 80.10 ± 1.16 & 81.92 ± 1.19 & 83.58 ± 1.09 & 84.68 ± 1.10 & 82.91 ± 1.25 & 84.25 ± 1.19 & 84.51 ± 1.16 & 84.81 ± 1.11 & 84.33 ± 1.16 \\
Aircraft & 52.97 ± 0.95 & 75.97 ± 0.96 & 77.01 ± 1.04 & 80.67 ± 0.99 & 83.88 ± 0.99 & 85.12 ± 0.99 & 85.95 ± 1.02 & 86.33 ± 0.95 & 85.45 ± 1.09 & 86.35 ± 0.95 & 85.04 ± 1.03 & 85.35 ± 0.99 & 83.44 ± 1.12 \\
 Birds & 83.12 ± 0.82 & 89.04 ± 0.69 & 90.40 ± 0.59 & 90.20 ± 0.67 & 90.92 ± 0.71 & 91.01 ± 0.74 & 91.22 ± 0.67 & 90.75 ± 0.75 & 90.50 ± 0.68 & 89.74 ± 0.77 & 89.17 ± 0.80 & 89.56 ± 0.74 & 88.63 ± 0.75 \\
 Textures & 84.95 ± 0.50 & 86.89 ± 0.58 & 88.34 ± 0.53 & 88.49 ± 0.52 & 88.25 ± 0.59 & 88.95 ± 0.56 & 88.52 ± 0.58 & 88.60 ± 0.51 & 87.83 ± 0.65 & 87.39 ± 0.64 & 86.20 ± 0.66 & 84.95 ± 0.71 & 85.47 ± 0.57 \\
 Quickdraw & 54.78 ± 0.94 & 63.47 ± 0.93 & 64.63 ± 0.96 & 67.80 ± 0.90 & 71.55 ± 0.94 & 73.69 ± 0.85 & 74.65 ± 0.85 & 75.29 ± 0.77 & 74.38 ± 0.75 & 75.55 ± 0.80 & 75.97 ± 0.73 & 75.49 ± 0.75 & 75.05 ± 0.83 \\
Fungi & 57.33 ± 1.06 & 61.30 ± 1.06 & 62.99 ± 1.18 & 63.86 ± 1.07 & 66.35 ± 1.08 & 66.87 ± 1.12 & 66.91 ± 1.06 & 66.64 ± 1.05 & 67.52 ± 1.05 & 65.03 ± 1.06 & 63.45 ± 1.10 & 63.59 ± 1.06 & 63.96 ± 1.08 \\
VGG\_Flower & 97.56 ± 0.25 & 97.10 ± 0.34 & 97.29 ± 0.33 & 97.27 ± 0.36 & 97.78 ± 0.30 & 98.06 ± 0.28 & 97.99 ± 0.29 & 97.88 ± 0.30 & 97.65 ± 0.31 & 97.55 ± 0.32 & 97.14 ± 0.36 & 97.00 ± 0.38 & 97.28 ± 0.32 \\ \hline
 Traffic Sign & 40.20 ± 1.07 & 52.42 ± 1.25 & 57.42 ± 1.27 & 60.77 ± 1.19 & 68.90 ± 1.23 & 75.15 ± 1.17 & 80.71 ± 1.17 & 85.52 ± 1.04 & 89.46 ± 0.91 & 91.29 ± 0.96 & 91.99 ± 0.96 & 92.76 ± 0.83 & 92.25 ± 0.96 \\
 MSCOCO & 54.13 ± 0.97 & 56.50 ± 0.95 & 58.64 ± 0.97 & 63.19 ± 1.01 & 63.22 ± 0.95 & 65.75 ± 1.00 & 65.59 ± 0.97 & 65.32 ± 0.93 & 64.70 ± 0.97 & 64.78 ± 0.95 & 63.46 ± 0.92 & 62.56 ± 1.00 & 62.02 ± 0.94 \\
 MNIST & 74.81 ± 0.74 & 86.77 ± 0.75 & 88.46 ± 0.73 & 89.86 ± 0.80 & 93.75 ± 0.68 & 94.51 ± 0.68 & 95.27 ± 0.65 & 96.13 ± 0.59 & 96.68 ± 0.53 & 96.87 ± 0.53 & 96.78 ± 0.50 & 97.51 ± 0.40 & 97.12 ± 0.45 \\
CIFAR-10 & 81.54 ± 0.64 & 86.54 ± 0.60 & 87.14 ± 0.65 & 87.60 ± 0.59 & 88.39 ± 0.58 & 89.17 ± 0.60 & 88.92 ± 0.61 & 89.04 ± 0.56 & 88.09 ± 0.61 & 87.40 ± 0.64 & 86.43 ± 0.68 & 85.75 ± 0.71 & 84.70 ± 0.76 \\
 CIFAR-100 & 73.41 ± 0.88 & 78.02 ± 0.79 & 78.57 ± 0.75 & 79.08 ± 0.78 & 81.32 ± 0.76 & 80.71 ± 0.76 & 80.99 ± 0.78 & 81.33 ± 0.81 & 81.45 ± 0.71 & 81.24 ± 0.78 & 80.02 ± 0.77 & 78.78 ± 0.79 & 78.53 ± 0.81 \\ \hline
Average Seen & 70.5	&76.8	&77.8	&79.4	&81.3	&82.1&	82.5&	\textbf{82.6} &82&	81.8&	81.2&	81&	80.7 \\
Average Unseen & 64.8	&72	&74	&76.1&	79.1&	81.1	&82.3&	83.5	&84.1	&\textbf{84.3}&	83.7	&83.5	&82.9 \\
Average All & 68.3&	75	&76.4	&78.1&	80.4	&81.7&	82.4&	\textbf{83}	&82.8	&82.8&	82.2&	82	&81.5 \\ \hline
\end{tabular}%
}
\caption{Variation of accuracies as the number of tuned layers $d_t$ varies in the MDL setting for in-domain and out-of-domain datasets in Meta-Dataset.}
\label{appendix_tab:tuning_depth_datasets}
\end{table*}

\subsection{Feature fusion depth}
\label{appendix:featuredepth}

We report the dataset-level accuracy values obtained as we vary the feature fusion depths in Supplementary Table \ref{appendix_tab:fusion-depth}, where a summary of it was presented in the main text.

\begin{table*}[!htbp]
\centering
\resizebox{\textwidth}{!}{%
\begin{tabular}{ccccccccc|cccccc}
\hline
Fusion depth & ImageNet             & Omniglot             & Aircraft             & Birds                & Textures             & Quick Draw           & Fungi                & VGG Flower           & Traffic Sign         & MS-COCO              & MNIST                & CIFAR-10             & CIFAR-100            & Average All      \\ \hline
1   & 70.2 ± 0.9          & 84.9 ± 1.1          & 86.2 ± 1.1          & 90.5 ± 0.7          & 88.3 ± 0.6          & 74.7 ± 0.8          & 66.5 ± 1.0          & 97.3 ± 0.4          & 90.3 ± 1.0          & 63.1 ± 0.9          & \textbf{97.4 ± 0.4} & 87.6 ± 0.6          & 80.4 ± 0.8          & 82.9           \\
2   & 70.5 ± 1.0          & 84.4 ± 1.2          & 86.7 ± 1.0          & 90.8 ± 0.7          & 87.9 ± 0.6          & 74.8 ± 0.8          & 66.6 ± 1.1          & 97.5 ± 0.3          & 90.4 ± 1.0          & 63.3 ± 1.0          & 97.0 ± 0.5          & \textbf{88.0 ± 0.6} & 80.0 ± 0.8          & 82.9           \\
4   & 70.9 ± 1.0          & 84.7 ± 1.1          & \textbf{86.3 ± 1.0} & 90.8 ± 0.8          & \textbf{88.6 ± 0.5} & \textbf{75.3 ± 0.8} & \textbf{66.6 ± 1.0} & \textbf{97.9 ± 0.3} & \textbf{91.3 ± 1.0} & \textbf{64.8 ± 1.0} & 96.9 ± 0.5          & 87.4 ± 0.6          & \textbf{81.2 ± 0.8} & \textbf{83..3} \\
6   & \textbf{71.6 ± 0.9} & \textbf{85.5 ± 1.1} & 86.1 ± 1.0          & \textbf{90.9 ± 0.7} & 88.0 ± 0.5          & \textbf{75.3 ± 0.8} & 66.0 ± 1.0          & 97.6 ± 0.4          & 90.3 ± 1.1          & 63.7 ± 1.0          & 97.3 ± 0.4          & 87.3 ± 0.6          & 80.8 ± 0.8          & 83.1           \\ 
8   & 69.2 ± 1.0          & 85.4 ± 1.1          & 85.6 ± 1.0          & 90.4 ± 0.8          & 87.4 ± 0.6          & 75.0 ± 0.8          & 65.4 ± 1.2          & 97.6 ± 0.4          & 90.7 ± 1.0          & 63.5 ± 1.0          & 96.9 ± 0.6          & 86.3 ± 0.7          & 80.0 ± 0.8          & 82.6           \\
12  & 68.0 ± 1.0          & 85.4 ± 1.1          & 85.7 ± 0.9          & 89.4 ± 0.9          & 87.7 ± 0.6          & 74.5 ± 0.8          & 63.6 ± 1.2          & 97.7 ± 0.3          & \textbf{91.3 ± 0.9} & 62.7 ± 1.0          & 97.2 ± 0.4          & 86.1 ± 0.6          & 78.6 ± 0.8          & 82.1       \\   \hline
\end{tabular}%
}
\caption{Variation of accuracies as the feature fusion depth $d_f$ vary on the MDL setting.}
\label{appendix_tab:fusion-depth}
\end{table*}

\subsection{Pre-training results}

The dataset-level accuracies reported in the SDL-E setting by DINO and MIM pre-trained models with varying fine-tuning strategies are reported in Supplementary Table \ref{appendix_tab:pre_training}.

\begin{table}[]
\centering
\resizebox{\columnwidth}{!}{%
\begin{tabular}{l|cc|cc}
\hline
 \multicolumn{1}{c|}{Pre-training} & \multicolumn{2}{c|}{MIM} & \multicolumn{2}{c}{DINO} \\ \cline{1-1} \hline
\multicolumn{1}{c|}{Fine-tuning} & NCC & DIPA & NCC & DIPA \\ \hline
ImageNet & 75.71 ± 0.81 & 77.26 ± 0.74 & 75.47 ± 0.82 & 75.89 ± 0.78 \\

\hline

Omniglot & 80.38 ± 1.36 & 84.06 ± 1.20 & 80.19 ± 1.31 & 83.65 ± 1.15 \\
Aircraft & 83.06 ± 1.03 & 87.09 ± 0.99 & 81.41 ± 1.10 & 85.88 ± 1.00 \\
Birds & 88.32 ± 0.75 & 90.52 ± 0.67 & 87.91 ± 0.77 & 90.37 ± 0.65 \\
Textures & 86.23 ± 0.69 & 87.32 ± 0.63 & 86.51 ± 0.72 & 87.06 ± 0.63 \\
Quickdraw & 73.38 ± 0.81 & 75.41 ± 0.81 & 72.62 ± 0.88 & 75.30 ± 0.79 \\
Fungi & 59.57 ± 1.08 & 60.89 ± 1.09 & 60.02 ± 1.14 & 62.16 ± 1.10 \\
VGG\_Flower & 96.88 ± 0.40 & 97.48 ± 0.36 & 96.56 ± 0.41 & 97.24 ± 0.35 \\
Traffic Sign & 89.93 ± 0.94 & 91.66 ± 0.84 & 89.68 ± 0.94 & 91.20 ± 0.81 \\
 MSCOCO & 64.52 ± 0.98 & 66.54 ± 0.93 & 64.30 ± 0.96 & 65.13 ± 0.99 \\
MNIST & 96.15 ± 0.50 & 97.24 ± 0.45 & 95.21 ± 0.63 & 96.82 ± 0.49 \\
CIFAR-10 & 90.23 ± 0.66 & 92.23 ± 0.47 & 88.20 ± 0.76 & 89.95 ± 0.66 \\
CIFAR-100 & 82.21 ± 0.79 & 84.48 ± 0.70 & 80.97 ± 0.76 & 82.29 ± 0.76 \\ \hline
Average Seen & 75.7 & \textbf{77.3} & 75.5 & 75.9 \\
Average Unseen & 82.6 & \textbf{84.6} & 82.0 & 83.9 \\
Average All & 82.0 & \textbf{84.0} & 81.5  & 83.3\\ \hline
\end{tabular}%
}
\caption{The impact of varying the pre-training and finetuning algorithms in SDL-E setting. }
\label{appendix_tab:pre_training}
\end{table}

\subsection{Further results on Meta-Dataset}

After evaluating our framework over a broad range of varying shots $K$ (e.g. up to 100 shots), we further analyze our framework in a more challenging setting. While $l_{A_\phi}$ requires at least two examples per class in order to gain benefits from its discriminative sample-based feature space adaptation, here we evaluate its performance in the more challenging varying-way, $5$-shot setting, comparing it with other works that have reported results in this context \cite{Li2021Cross-domainAdapters}. As shown in Supplementary Table \ref{appendix_tab:MD_VARYING_5_SHOT}, overall performance for all methods has decreased due to the even more challenging nature of the support set. Nevertheless, our method still outperforms the existing methods when the number of support images per class is fewer, especially on the challenging unseen domains by 2.9\%.



\begin{table}[]
\centering
\resizebox{\columnwidth}{!}{%
\begin{tabular}{l|ccccc}
\hline
 & \begin{tabular}[c]{@{}c@{}}Simple\\ CNAPS\end{tabular} & SUR & URT & TSA & DIPA \\ \hline
 SS PT & & & & & $\checkmark$ \\
 Sup. MT & $\checkmark$&$\checkmark$ &$\checkmark$ &$\checkmark$ &  \\
 Backbone &RN18 & RN18& RN18& RN18& ViT-s \\ \hline
ImageNet & 47.2 ± 1.0 & 46.7 ± 1.0 & 48.6 ± 1.0 & 48.3 ± 1.0 & \textbf{60.17 ± 0.80} \\
Omniglot & 95.1 ± 0.3 & 95.8 ± 0.3 & 96.0 ± 0.3 & \textbf{96.8 ± 0.3} & 91.30 ± 0.46 \\
Aircraft & 74.6 ± 0.6 & 82.1 ± 0.6 & 81.2 ± 0.6 & \textbf{85.5 ± 0.5} & 64.77 ± 0.68 \\
Birds & 69.6 ± 0.7 & 62.8 ± 0.9 & 71.2 ± 0.7 & 76.6 ± 0.6 & \textbf{87.55 ± 0.39} \\
Textures & 57.5 ± 0.7 & 60.2 ± 0.7 & 65.2 ± 0.7 & 68.3 ± 0.7 & \textbf{79.69 ± 0.50} \\
Quickdraw & 70.9 ± 0.6 & 79.0 ± 0.5 & \textbf{79.2 ± 0.5} & 77.9 ± 0.6 & 68.40 ± 0.68 \\
Fungi & 50.3 ± 1.0 & 66.5 ± 0.8 & 66.9 ± 0.9 & \textbf{70.4 ± 0.8} & 66.57 ± 0.77 \\
VGG\_Flower & 86.5 ± 0.4 & 76.9 ± 0.6 & 82.4 ± 0.5 & 89.5 ± 0.4 & \textbf{96.96 ± 0.18} \\ \hline
Traffic Sign & 55.2 ± 0.8 & 44.9 ± 0.9 & 45.1 ± 0.9 & 72.3 ± 0.6 & \textbf{83.91 ± 0.45} \\
MSCOCO & 49.2 ± 0.8 & 48.1 ± 0.9 & 52.3 ± 0.9 & 56.0 ± 0.8 & \textbf{64.64 ± 0.68} \\
MNIST & 88.9 ± 0.4 & 90.1 ± 0.4 & 86.5 ± 0.5 & \textbf{92.5 ± 0.4} & 92.07 ± 0.33 \\
CIFAR-10 & 66.1 ± 0.7 & 50.3 ± 1.0 & 61.4 ± 0.7 & 72.0 ± 0.7 & \textbf{80.37 ± 0.53} \\
CIFAR-100 & 53.8 ± 0.9 & 46.4 ± 0.9 & 52.5 ± 0.9 & 64.1 ± 0.8 & \textbf{76.79 ± 0.64} \\ \hline
Average Seen & 69.0 & 71.2 & 73.8 & 76.7 & \textbf{76.9} \textcolor{green}{(+0.2)} \\
Average Unseen & 62.6 & 56.0 & 59.6 & 71.4 & \textbf{74.3} \textcolor{green}{(+2.9)} \\
Average All & 66.5 & 65.4 & 68.3 & 74.6 & \textbf{76.4} \textcolor{green}{(+2.2)} \\ \hline
\end{tabular}%
}
\caption{Results of Varying-Way Five-Shot in the MDL setting. Average (Avg.) accuracies are reported. RN: ResNet, ViT-s: ViT-small, SS PT: indicates self-supervised pre-training and Sup. MT: indicates supervised meta-training. 
}
\label{appendix_tab:MD_VARYING_5_SHOT}
\end{table}




\section{Addtional Results for  miniImageNet and CIFAR-FS}

Supplementary Table  \ref{appendix_tab:5-way-5-shot} reports the results for evaluating the DIPA framework under the SDL-E setting on CIFAR-FS and mini-ImageNet datasets. Here, we follow PMF \cite{Hu2022PushingDifference} and compare DIPA with relevant existing methods. Our approach can be directly compared with methods that employ SSL for pre-training, both with and without subsequent fine-tuning (Supplementary Table \ref{appendix_tab:5-way-5-shot}, row D0-D4). Among those methods, our approach has superior performance across most scenarios. Notably, among the other methods that use various other training strategies, we still obtain somewhat good performance without requiring additional meta-training or training labels.

\begin{table}[!htbp]
\centering
\resizebox{\columnwidth}{!}{%
\begin{tabular}{ll|c|cc|cc|cc}
\hline
ID & Method & Backbone & Ext. & Ext. & \multicolumn{2}{c|}{miniImageNet} & \multicolumn{2}{c}{CIFAR-FS} \\ \cline{6-9} 
& &  & dat. & lab. & 5/1 & 5/5 & 5/1 & 5/5 \\ \hline

 & \textbf{Inductive} &  &  &  &  &  &  &  \\
A0 & Baseline++ \cite{wei2019closerlook} & CNN-4-64 &  &  & 48.2 ± 0.8 & 66.4 ± 0.6 &  &  \\
A1 & MetaOpt-SVM \cite{lee2019meta} & ResNet12 &  &  & 62.6 ± 0.6 & 78.6 ± 0.5 & 72.0 ± 0.7 & 84.2 ± 0.5 \\
A2 & Meta-Baseline \cite{chen2021meta} & ResNet12 &  &  & 63.2 ± 0.2 & 79.3 ± 0.2 &  &  \\
A3 & RS-FSL \cite{afham2021} & ResNet12 &  & $\checkmark$ & 65.3 ± 0.8 &  &  &  \\
& \textbf{Transductive} &  &  &  &  &  &  &  \\
B0 & Fine-tuning \cite{dhillon2020} & WRN-28-10 &  &  & 65.7 ± 0.7 & 78.4 ± 0.5 & 76.6 ± 0.7 & 85.8 ± 0.50 \\
B1 & SIB \cite{shell2020} & WRN-28-10 &  &  & 70.0 ± 0.6 & 79.2 ± 0.4 & 80.0 ± 0.6 & 85.3 ± 0.4 \\
B2 & PT-MAP \cite{hu2021leveraging} & WRN-28-10 &  &  & 82.9 ± 0.3 & 88.8 ± 0.1 & \textbf{87.7 ± 0.2}$^\bullet$ & 90.7 ± 0.2 \\
B3 & CNAPS + FETI \cite{bateni2022enhancing} & WRN-28-10 & $\checkmark$ & $\checkmark$ & 79.9 ± 0.8 & 91.5 ± 0.4 &  &  \\
 & \textbf{Semi-Supervised} &  &  &  &  &  &  &  \\
C0 & LST \cite{li2019learning} & ResNet12 & $\checkmark$ &  & 70.1 ± 1.9 & 78.7 ± 0.8 &  &  \\
C1 & PLCM \cite{huang2021pseudo} & ResNet12 & $\checkmark$ &  & 72.1 ± 1.1 & 83.7 ± 0.6 & 77.6 ± 1.2 & 86.1 ± 0.7 \\ 

& \textbf{Self-Supervised} &  &  &  &  &  &  &  \\
D0 & ProtoNet \cite{gidaris2019boosting} & WRN-28-10 &  &  & 62.9 ± 0.5 & 79.9 ± 0.3 & 73.6 ± 0.3 & 86.1 ± 0.2 \\
D1 & ProtoNet \cite{chen2021self} & AMDIM ResNet & $\checkmark$ &  & 76.8 ± 0.2 & 91.0 ± 0.1 &  &  \\
D2 & EPNet + SSL \cite{rodriguez2020embedding} & WRN-28-10 & $\checkmark$ &  & 79.2 ± 0.9 & 88.1 ± 0.5 &  &  \\
D3 & FewTure \cite{Hiller2022RethinkingClassification} & ViT-small &  &  & 68.0 ± 0.9 & 84.5 ± 0.5 & \textbf{76.1 ± 0.9}* & 86.1 ± 0.6 \\
D4 & DIPA & ViT-small & $\checkmark$ &  & \textbf{79.6 ± 0.7}* & \textbf{94.3 ± 0.3}* & 65.2 ± 0.9 & \textbf{88.4 ± 0.6}* \\ 
 & \textbf{Self-Supervised + MT} &  &  &  &  &  &  &  \\
E0 & PMF \cite{Hu2022PushingDifference} & ViT-small & $\checkmark$ &  & \textbf{93.1}$^\bullet$ & \textbf{98.0}$^\bullet$ & 81.1 & \textbf{92.5}$^\bullet$ \\ 
\hline
\end{tabular}%
}
\caption{Comparison with representative state-of-the-art FSL algorithms on miniImageNet \& CIFAR-FS for 5-way-1-shot (5/1) and 5-way-5-shot (5/5). Mean accuracy and 95\% confidence interval are reported, where available. $\checkmark$ indicates the use of Extra data or Extra labels. MT: Meta-train, and * denotes the highest performance among the most relevant methods that are directly comparable to DIPA while $^\bullet$ denotes the highest performance overall.} 
\label{appendix_tab:5-way-5-shot}
\end{table}